\documentclass{article}
\usepackage{iclr2027_conference,times}

\usepackage{amsmath,amsfonts,bm}

\def\eqref#1{equation~\ref{#1}}

\def\1{\bm{1}}

\DeclareMathAlphabet{\mathsfit}{\encodingdefault}{\sfdefault}{m}{sl}
\SetMathAlphabet{\mathsfit}{bold}{\encodingdefault}{\sfdefault}{bx}{n}

\usepackage{amsmath}
\usepackage{array}
\usepackage{booktabs}
\usepackage{float}
\usepackage{placeins}
\usepackage[ruled,lined]{algorithm2e}
\usepackage{graphicx}
\usepackage{xcolor}
\usepackage{colortbl}
\usepackage{url}

\definecolor{linkblue}{RGB}{0,76,130}
\definecolor{refred}{RGB}{255,0,0}
\definecolor{citeteal}{RGB}{108, 150, 188}
\usepackage[
  colorlinks=true,
  linkcolor=refred,
  citecolor=citeteal,
  urlcolor=linkblue,
  pdfborder={0 0 0}
]{hyperref}

\newcommand{\more}{{\upshape MORE}}
\newcommand{\moremodel}{\textsc{MORE}}
\newcommand{\mode}{a_{\mathrm{mode}}}
\newcommand{\modeset}{\mathcal{A}_{\mathrm{mode}}}
\newcommand{\rmore}{R_{\mathrm{MORE}}}
\newcommand{\moremethod}{\textbf{M}ode-\textbf{O}ptimized \textbf{R}easoning with Self-Adaptive \textbf{E}ffort}
\newcommand{\humanevalplus}{HumanEval\textsuperscript{+}}
\newcommand{\heplus}{HE\textsuperscript{+}}
\newcommand{\mbppplus}{MBPP\textsuperscript{+}}
\newcommand{\tablesource}[1]{\textsuperscript{[\citenum{#1}]}}
\definecolor{moregreen}{HTML}{2E9E44}
\definecolor{moreblue}{HTML}{DCE6FF}
\definecolor{tablegroupgray}{HTML}{F5F5F5}
\newcommand{\moregain}[1]{\textsuperscript{\textcolor{moregreen}{\scriptsize #1}}}
\definecolor{rqpurple}{HTML}{F0E8F7}
\newcommand{\rqbox}[2]{%
  \par\noindent\colorbox{rqpurple}{%
    \parbox{\dimexpr\linewidth-2\fboxsep\relax}{\textit{\textbf{RQ#1:} #2}}%
  }\par\vspace{0.65em}%
}
\newcommand{\appendixtocsection}[2]{\noindent\hyperref[#1]{\textcolor[HTML]{C9A5DE}{\textbf{#2}}}\par\vspace{0.28em}}
\newcommand{\appendixtocsubsection}[2]{\noindent\hspace*{1.2em}\hyperref[#1]{\textcolor[HTML]{C9A5DE}{#2}}\par\vspace{0.14em}}

\title{MORE: Mode-Optimized Reasoning with Self-Adaptive Effort for Self-Evolving Agents}
\author{
Dianxing Shi$^{1}$, Dingjie Fu$^{2}$, Tianyu Lei$^{1}$, Wangjie Qiu$^{1}$\thanks{Corresponding author} \\
\\
$^{1}$Beihang University, $^{2}$Huazhong University of Science and Technology \\
\{dianxingshi, wangjieqiu\} @buaa.edu.cn
}    

\iclrfinalcopy
\begin{document}

\maketitle

\begin{abstract}
Self-evolving training improves the performance of agents on difficult tasks through verifiable feedback, but it often applies costly deliberation uniformly across tasks. To address this problem, we propose Mode-Optimized Reasoning with Self-Adaptive Effort (MORE), a self-evolving training framework that adaptively selects both reasoning form and computational effort for each task. Specifically, MORE follows a two-stage training paradigm: (i) \emph{Stage 1} establishes a compact prior over direct answering, short and long chain-of-thought, and code-based reasoning, and (ii) \emph{Stage 2} self-evolves this prior with Mode-Adaptive and Test-time Effort PPO (MATE-PPO) on generated, verifiable task streams without curated data. By learning mode selection and token expenditure jointly, MORE allocates concise reasoning to tasks that permit it while retaining extended deliberation where it is useful. Across coding, mathematics, and general-reasoning benchmarks, MORE achieves consistent performance gains on Base and Coder backbones at model scales from 3B to 14B. During inference, MORE demonstrates higher accuracy and better efficiency, with markedly shorter response lengths and reduced inference time.
\end{abstract}
\begin{figure*}[htbp]
  \centering
  \includegraphics[width=0.97\columnwidth]{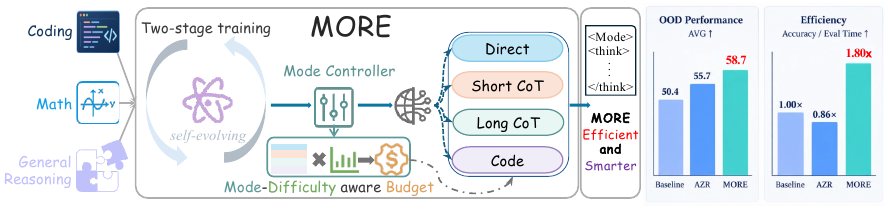}
  \vspace{-0.36em}
  \caption{\textbf{Left}: \more{} uses Two-Stage self-evolving training with a mode-difficulty aware budget to select the most efficient reasoning mode for each task while maintaining appreciable accuracy. \textbf{Right}: Across coding, mathematics, and general-reasoning tasks, \more{} markedly improves average accuracy and efficiency over both its underlying backbone and Absolute-Zero Reasoner (AZR).}
  \label{figs1}
\end{figure*}
\section{Introduction}
Self-evolving training, especially for coding and mathematics reasoning, is a post-training approach that aims to strengthen agents' reasoning ability without human-supervision. ‌The prevailing existing methods designed for coding and mathematics reasoning use Proposer-Solver paradigm for reasoning models in which the model simultaneously learns to define tasks that maximize learnability and to solve them effectively, enabling self-evolution through self-play without relying on external data. In practice, people find that reasoning-oriented language models improve difficult problem solving by making intermediate computation explicit, showcasing superior performance.

Typically, this intermediate computation is mainly produced by Chain-of-thought and Reinforcement learning. 
Chain-of-thought prompting and self-consistency establish that generated reasoning can improve answer quality \citep{wei2023chainofthoughtpromptingelicitsreasoning,wang2023selfconsistencyimproveschainthought}, while test-time scaling shows that additional sampling or search can convert more inference compute into higher accuracy \citep{brown2024largelanguagemonkeys,snell2024scalingllmtesttimecompute}. Reinforcement learning with verifiable rewards extends this principle into training: execution, answer extraction, and formal checks provide scalable feedback for coding and mathematical reasoning \citep{deepseekai2026deepseekr1incentivizingreasoningcapability,zhao2025absolutezeroreinforcedselfplay}. Despite their successes, these objectives do not specify how much computation is appropriate for each task.

To this end, the absence of accuracy-computation calibration becomes consequential in self-evolving agents. Generated tasks, sampled solutions, and verifier outcomes continually reshape both capability and effective task difficulty. Repeated exposure to difficult verifiable tasks can make long reasoning a default hedge, even when a direct answer would suffice, increasing response length and completion time. Autonomous evolution can also introduce capability degradation and safety drift across coding, mathematical, and safety evaluations \citep{shi2026towardshealthyevolution}. Long reasoning is valuable when needed, but self-evolving agents must allocate it where its accuracy gains justify its cost. 


Difficulty-aware compute allocation addresses only part of this problem. CODA, for example, argues that simple instances often gain little from additional tokens whereas hard instances may still require deliberate computation \citep{wu2026codadifficultyawarecomputeallocation}. A scalar budget alone, however, can not decide whether the computation should be a direct response, a long derivation, or executable code. Conversely, choosing a discrete reasoning format without controlling its length can still produce verbose traces. Self-evolution adds another complication: a static reward schedule is calibrated to one policy state, even though the solver's success profile and useful budget change during training.

Given the above, we propose \moremethod{} (\more), a self-evolving training framework that learns four unique reasoning forms and computation amount as coupled policy decisions. Specifically, for task $x$ and response $y$, \more{} factorizes generation as $\pi_\theta(\mode,y\mid x)=\pi_\theta(\mode\mid x)\pi_\theta(y\mid x,\mode)$, where $\mode$ is one of \texttt{direct}, \texttt{short\_cot}, \texttt{long\_cot}, or \texttt{code}. This factorization separates the form of computation from its amount while allowing both decisions to be optimized by the same training signal. Before producing the response, \more{} emits a structured \texttt{<mode>} control token as its first internal decision. Thereafter, the token conditions the subsequent response and supports training and inference statistics.

\more{} learns this behavior in two stages. Mode-format supervision first makes the action vocabulary identifiable and gives each mode a stable response contract. \textbf{M}ode-\textbf{A}daptive \& \textbf{T}est-time \textbf{E}ffort PPO (MATE-PPO) is our proposed Stage-2 optimization method for coupled mode--budget learning \citep{schulman2017proximalpolicyoptimization}. It integrates verifier-grounded correctness, online difficulty estimation, mode fit, and dynamic budget regulation into a single trajectory objective, so clipped policy updates jointly shape reasoning form and computation amount as a learned model capability.

\textbf{Our core contributions are as follows:} \textbf{i)} We introduce \more, a self-evolving training framework that jointly adapts reasoning mode and computation budget to online task difficulty. Across coding, mathematical, and general-reasoning benchmarks, \more{} delivers strong and consistent performance. \textbf{ii)} Our two-stage learning design first establishes four mode-specific response contracts and then applies MATE-PPO to jointly optimize mode choice, response generation, and computational effort through a unified verifier-grounded trajectory objective. \textbf{iii)} We design an online difficulty--effort controller that converts evolving rollout clusters into difficulty tiers and updates mode-conditioned budget bands throughout training, enabling the policy to calibrate its reasoning allocation as its capabilities change. \more{} also reaches its strongest observed out-of-distribution performance earlier during self-evolution than another framework in the same self-evolving paradigm.

\section{Related Work}


\textbf{Mode-aware reasoning.}
Hybrid reasoning models expose multiple response forms so that a solver need not apply one deliberation pattern to every query. Representative approaches distinguish direct answers from longer reasoning or combine textual and code-mediated computation \citep{wu2025armadaptivereasoningmodel,luo2025adar1hybridcotbileveladaptive}. Uniform response behavior can be brittle under task-specific requirements \citep{fu2025rejected}. A fixed mode policy, however, does not track the changing competence of a self-evolving solver, and a mode choice alone does not constrain token use within that mode.

\textbf{Adaptive reasoning budgets.}
Inference-time scaling, budget forcing, and adaptive length objectives control how much computation a model spends \citep{muennighoff2025s1simpletesttimescaling,zhu2025scalingtesttimecomputeagents}. LASER conditions length-based reward shaping on difficulty \citep{liu2025learnreasonefficientlyadaptive}, and further, CODA jointly studies difficulty estimation and compute allocation \citep{wu2026codadifficultyawarecomputeallocation}. This line demonstrates that uniform token budgets are inefficient, but it largely treats response format as fixed or allocates compute after the mode of reasoning has already been chosen. \more{} addresses the intersection: it trains a self-evolving model based on the Proposer-Solver paradigm with a shared policy over reasoning mode and budget, and updates the associated control state as capability changes.

\section{Methodology}

\subsection{\more{} Framework}

Our proposed Mode-Optimized Reasoning with Self-Adaptive Effort (MORE) is a two-stage self-evolving training framework that enables models to learn both the form and the amount of computation used for each task. The model trained under MORE framework first emits one of four internal modes: \texttt{direct}, \texttt{short\_cot}, \texttt{long\_cot}, or \texttt{code}, and then generates a response under the corresponding reasoning contract and a difficulty-conditioned effort budget.

\begin{figure*}[t]
  \centering
  \includegraphics[width=\columnwidth]{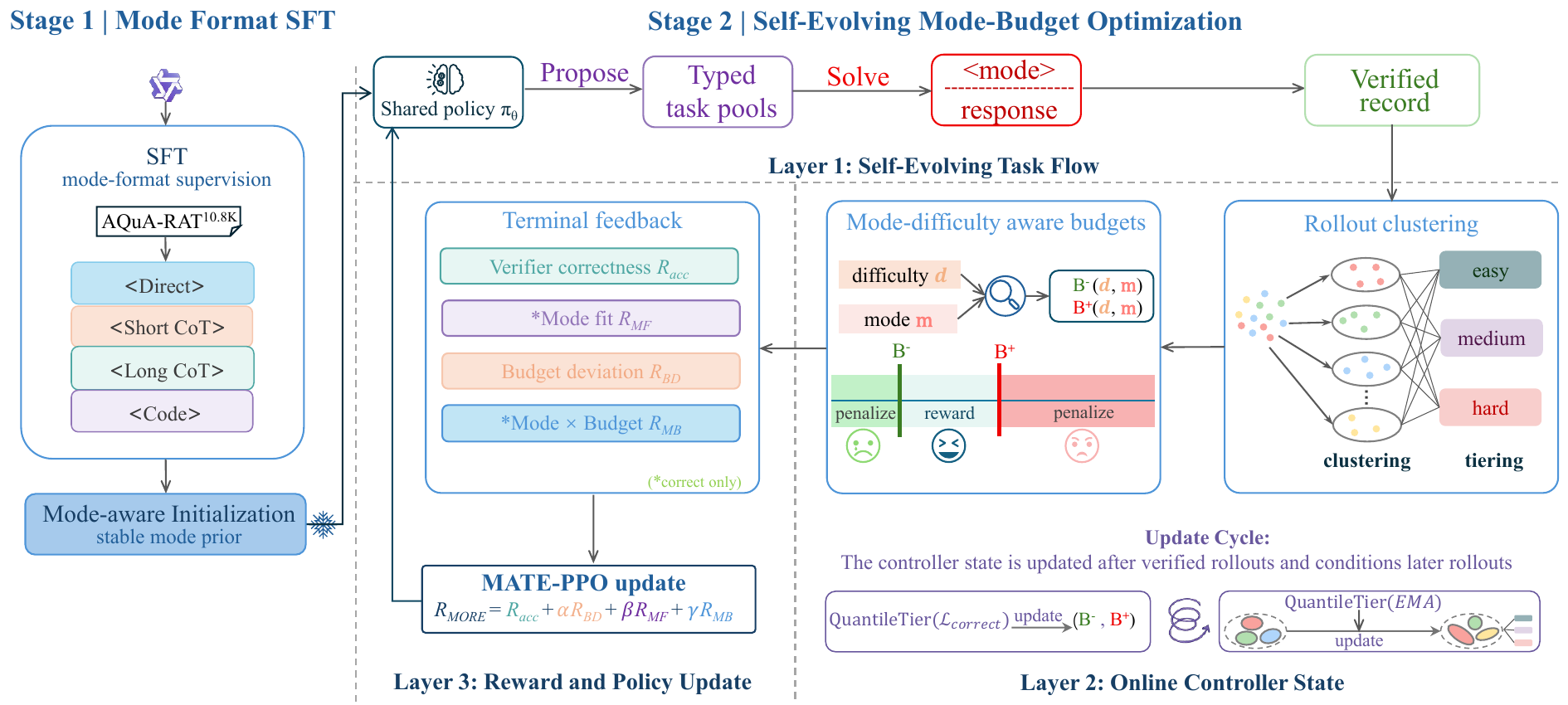}
  \caption{Overview of the \more{} framework. Mode-format supervision initializes four reasoning contracts. MATE-PPO then couples self-evolving verifiable rollouts with online difficulty estimation, mode--budget control, a joint terminal return, and a clipped policy update. The diagram summarizes correctness gating: mode-fit and joint credit require a correct response, while the budget-deviation term regulates overlength for both correct and incorrect responses.}
  \label{fig:method-pipeline}
\end{figure*}

Figure~\ref{fig:method-pipeline} summarizes the complete MORE framework. MORE's Stage 1 constructs mode-format supervision to initialize a stable prior over the four reasoning contracts. \more{}'s Stage 2 draws inspiration from AZR's proposer--solver self-evolving paradigm \citep{zhao2025absolutezeroreinforcedselfplay} and it performs the overall reinforcement-learning procedure: it samples verifiable tasks, generates mode-conditioned rollouts, verifies their outcomes, updates online difficulty and budget state, constructs the \more{} terminal return, and applies a Mode-Adaptive \& Test-time Effort (MATE)
PPO update. Telemetry from the same loop records capability, mode allocation, token economy, and system efficiency.

Stage 2 operationalizes \more{} through a self-evolving proposer--solver training loop, with MATE-PPO providing its policy-optimization mechanism. For each sampled task, \more{} generates a mode token and response under a pre-update difficulty context and mode-conditioned budget band. The verifier then supplies correctness, which MATE-PPO combines with the retained controller state to form the terminal return for a clipped policy update. The controller then updates the associated cluster's difficulty context and mode-conditioned budget band for subsequent rollouts. The generated mode token remains a likelihood-bearing part of the optimized trajectory, linking mode selection, response generation, and effort calibration. Appendix~\ref{app:stage2-training} gives the complete training process.

\subsection{Stage 1: Mode-Format Initialization}

First of all, Stage 1 builds a response-format prior before sparse outcome feedback is introduced. Specifically, each question is paired with one of four output contracts: an answer-only record for \texttt{direct}, a compact \texttt{COT} rationale for \texttt{short\_cot}, a detailed and complex \texttt{COT} rationale for \texttt{long\_cot}, or executable \texttt{CODE} followed by an answer for \texttt{code}. The tags serve as internal supervision targets and do not appear as user-facing method names. They make response form observable to the model while preserving a common final-answer field to achieve consistent universality.

For training data, all backbones use the same SFT corpus, \mbox{AQuA-RAT}\textsuperscript{10.8K}, obtained by processing the original AQuA-RAT examples into four reasoning-mode formats \citep{ling2017programinductionrationalegeneration,wu2025armadaptivereasoningmodel}. Appendix~\ref{app:sft-datasets} gives the complete construction details. The resulting tuples $(x,\mode,y)$ are optimized with $\max_\theta\mathbb{E}_{\mathcal D_{\mathrm{sft}}}[\log\pi_\theta(\mode,y\mid x)]$. Stage 1 therefore establishes the mode vocabulary, response contracts, and the prior needed for autonomous mode-token generation in Stage 2.

\subsection{Stage 2: Self-Evolving Mode--Budget Optimization}

We implement the Stage-2 update with MATE-PPO by treating the generated mode token and its subsequent response as one policy trajectory. The verifier-derived correctness score, online difficulty state, mode-fit utility, and budget-deviation signals jointly define the trajectory return. With likelihood ratio $r_t(\theta)=\pi_\theta(a_t\mid s_t)/\pi_{\mathrm{old}}(a_t\mid s_t)$, MATE-PPO aims to maximize
\begin{equation}
  \mathcal L_{\mathrm{MATE\text{-}PPO}}(\theta)=
  \mathbb E_t\!\left[\min\!\left(r_t\hat A_t^{\mathrm{MORE}},
  \operatorname{clip}(r_t,1-\epsilon,1+\epsilon)\hat A_t^{\mathrm{MORE}}\right)\right].
  \label{eq:mate-ppo}
\end{equation}
Here, MATE-PPO constructs $\hat A^{\mathrm{MORE}}$ from the mode-budget-aware terminal return and controller state, then optimizes the joint trajectory with a clipped policy objective. The likelihood-bearing mode token selects the associated mode utility and budget band, and contributes to mode ratios, per-mode success and mode-difficulty telemetry. The following subsections provide further details.

\subsection{Adaptive Mode-Budget Control}

\paragraph{Online difficulty.}
The controller represents each task by structural and behavioral features. Firstly, Structural features include code shape, prompt artifacts, task family, and verifier target; behavioral features include rolling success and response length. Thus, a global success threshold would become stale as a self-evolving policy improves, whereas independent per-task estimates are sparse and noisy. Then,  Clustering pools evidence across structurally and semantically comparable rollouts; the EMA follows policy drift without reacting fully to one outcome; and quantile tiers preserve relative difficulty when absolute success rises. Finally, the resulting tier indexes both mode-fit utility and mode-conditioned budget bands, applying stronger pressure toward direct or short traces in currently easy regions while retaining wider long-reasoning or code allowances in hard regions.

Subsequently, a stable task identifier is assigned to a family- and verifier-aware cluster $c$. After a verified rollout, its success estimate is updated by $q_c^{(t)}=\rho q_c^{(t-1)}+(1-\rho)c_t$, where $c_t\in\{0,1\}$ and $\rho=0.8$. Mature clusters are ranked by success: clusters above the $0.67$ quantile are easy, those at or below the $0.33$ quantile are hard, and the remainder are medium. The updated state is carried into subsequent rollouts, allowing mode preferences and budget bands to track the evolving policy rather than a fixed notion of task difficulty. The complete rollout representation, family and verifier partitioning, centroid and EMA updates, among other details, are provided in Appendix~\ref{app:rollout-clustering}.

\paragraph{Autonomous mode-token generation.}
For every rollout, \more{} generates a \texttt{<mode=m>} control token before the response body, with $m\in\{\text{direct},\text{short-CoT},\text{long-CoT},\text{code}\}$. This token is the realized $a_{\mathrm{mode}}$: it selects the response contract, indexes $U(d,m)$, and identifies the mode-conditioned budget band. As a likelihood-bearing policy action, it makes mode allocation an endogenous model behavior under the task distribution.

\paragraph{Mode-conditioned budget bands.}
Each $(d,m)$ cell has a lower target $B^-_{d,m}$ and upper target $B^+_{d,m}$. Correct response lengths are accumulated through hierarchical keys, from cluster/family/verifier/mode statistics to tier/mode aggregates. Once enough observations are available, the lower and upper targets use empirical quantiles $Q_{0.2}$ and $Q_{0.8}$; sparse cells retain the initialized prior. Downward updates are rate-limited and the lower target is constrained below the upper target, preventing one short batch from collapsing the band. The bands provide a soft learning signal that preserves flexible generation while progressively calibrating token use and  the inference-time latent budget is determined by the band at final training checkpoint. Appendix~\ref{app:reward-defaults} reports their initialization.

\subsection{The \more{} Terminal Return}

The return used by MATE-PPO is
\begin{equation}
  \rmore=R_{\mathrm{acc}}+\alpha R_{\mathrm{BD}}+\beta R_{\mathrm{MF}}+\gamma R_{\mathrm{MB}},
  \label{eq:more-return}
\end{equation}
where $R_{\mathrm{BD}}$ is the budget-deviation term, $R_{\mathrm{MF}}$ is the mode-fit term, and $R_{\mathrm{MB}}$ is the mode--budget joint term. For solver trajectories, the runtime composes verifier-provided $R_{\mathrm{acc}}$ with $\Delta_{\mathrm{MORE}}=\alpha R_{\mathrm{BD}}+\beta R_{\mathrm{MF}}+\gamma R_{\mathrm{MB}}$, realizing the complete terminal return while counting execution feedback exactly once. Appendix~\ref{app:mate-algorithm} gives the parallel role-specific form for all proposer--solver streams.

\paragraph{Budget-deviation reward.}
For response length $L$, define normalized deviations $o=\max(0,(L-B^+_{d,m})/B^+_{d,m})$ and $u=\max(0,(B^-_{d,m}-L)/B^-_{d,m})$. The main realization uses $R_{\mathrm{BD}}=\operatorname{clip}[-\lambda_d(w_o o+w_u(c)u),-0.5,0]$, with $w_o=1$, $w_u(1)=1$ for correct responses, and $w_u(0)=0.1$ for incorrect responses. It penalizes overlong responses regardless of correctness, down-weights lower-bound pressure on incorrect responses, and ensures that brevity earns efficiency credit only together with correctness.

\paragraph{Mode-fit reward.}
For a correct response, $R_{\mathrm{MF}}=U(d,m)$. The structured utility prior favors direct or short reasoning for easy tasks and allocates stronger credit to long reasoning or code as difficulty rises. Stage 1 makes these formats identifiable; $R_{\mathrm{MF}}$ then supplies outcome-conditioned credit for selecting them appropriately.

\paragraph{Mode--budget joint term.}
For correct responses, $R_{\mathrm{MB}}=\operatorname{clip}[-\eta(d,m)(o+u),-0.25,0]$. This coupling directly measures mode-specific budget deviation and supplies a bounded joint learning signal. The main run uses $(\alpha,\beta,\gamma)=(0.6,1.0,0.05)$.

\subsection{Inference and Telemetry Interface}

The same four-mode vocabulary is used in supervision, rollout generation, reward routing, inference, and evaluation. \more{} emits the mode token before the response body and then follows the corresponding response contract. The inference harness parses the generated token for telemetry and scores the resulting answer. The telemetry interface records mode ratios, per-mode success, difficulty tiers, budget bands, reward components, response-length distributions, throughput, and wall-clock components, providing a unified view of adaptive reasoning behavior.

\section{Experiments Setup}

\subsection{Models and Training Configuration}

We use Qwen2.5-Coder-7B and Qwen2.5-7B-Base as backbones \citep{qwen2025qwen25technicalreport}. Additional experiments include Qwen2.5-Coder-3B, Qwen2.5-Coder-14B, and Qwen2.5-14B-Base \citep{hui2024qwen25coder,qwen2025qwen25technicalreport}. The complete training configuration and Stage-1 data construction are provided in Appendices~\ref{app:training-metrics} and~\ref{app:sft-datasets}, respectively. Also, Appendix~\ref{app:stage2-training} and Table~\ref{tab:training-hyperparameters} specify the Stage-2 rollout structure, optimization settings, checkpoint cadence, and controlled-ablation protocol.

\subsection{Capability Evaluation and Efficiency Metrics}

We use CruxEval-I/O \citep{gu2024cruxeval} and LiveCodeBench-Execution \citep{jain2024livecodebench} as our in-distribution (ID) capability benchmarks. Further, we group the out-of-distribution (OOD) capability benchmarks into code and mathematical reasoning: Code evaluation uses \humanevalplus{}, \mbppplus{}, and LiveCodeBench Generation v1--v5 \citep{liu2023evalplus,jain2024livecodebench}. Mathematical evaluation uses AIME 2024, AIME 2025, AMC 2023, MATH-500, Minerva Math, and OlympiadBench \citep{hendrycks2021math,he2024olympiadbench,lewkowycz2022solvingquantitative}. We additionally report GSM8K and MMLU-Pro as supplemental general-reasoning benchmarks. Appendix~\ref{app:training-metrics} gives metric definitions and decoding settings. \textsc{MORE}-Auto uses mode-neutral prompts: it emits \texttt{<mode>} before answering, and the harness records that token without modifying the response.

We assess efficiency during both self-evolving training and out-of-distribution testing through computational demand, response behavior, elapsed time, processing throughput, and convergence. For out-of-distribution evaluation, \textbf{Response Length Avg} is the sample-mean number of generated response tokens, and \textbf{Wall Time} is the total elapsed time for the corresponding benchmark suite. Under matched software, hardware, and decoding settings, we track token consumption, Response Length Avg, Wall Time, mode--budget allocation, and GPU-normalized throughput. Relating these measurements to capability trajectories over optimizer steps and cumulative training tokens shows how quickly each method reaches strong performance and at what computational cost.

\section{Experimental Results}

\subsection{Main Results}
\label{main results}
\begin{table*}[!t]
\caption{Performance of reasoning models based on Qwen2.5-7B variants. \textbf{CAvg} is the mean of \humanevalplus{}, \mbppplus{}, and the LiveCodeBench v1--v5 average; \textbf{MAvg} is the mean of the six mathematical reasoning benchmarks. \textbf{AVG} is the mean of CAvg, MAvg, GSM8K, and MMLU-Pro. A24/A25 denote AIME 2024/2025, M500 denotes MATH-500, Min. denotes Minerva Math, and Oly. denotes OlympiadBench. The best and second-best results are shown in \textbf{bold} and \underline{underlined}, respectively. Appendix~\ref{app:backbone-provenance} provides more implementation, training-data, and initialization details.}
\label{tab:main-results}
\centering
\small
\setlength{\tabcolsep}{1.0pt}
\renewcommand{\arraystretch}{1.04}
\resizebox{\textwidth}{!}{%
\begin{tabular}{@{}l|*{4}{c}|*{7}{c}|*{2}{c}|>{\columncolor{black!8}}c@{}}
\toprule
\textbf{Model} & \textbf{\heplus} & \textbf{\mbppplus} & \textbf{LCB} & \textbf{CAvg} & \textbf{A24} & \textbf{A25} & \textbf{AMC} & \textbf{M500} & \textbf{Min.} & \textbf{Oly.} & \textbf{MAvg} & \textbf{GSM8K} & \textbf{M-Pro} & \textbf{AVG} \\
\midrule
\rowcolor{tablegroupgray}\multicolumn{15}{c}{\textbf{Base Models}} \\
\midrule
Qwen2.5-7B-Base\tablesource{qwen2025qwen25technicalreport} & 73.2 & 65.3 & 17.5 & 52.0 & 6.7 & 3.3 & 37.5 & 64.8 & 25.0 & 27.7 & 27.5 & 80.4 & 47.6 & 51.9 \\
Qwen2.5-Coder-7B\tablesource{hui2024qwen25coder} & 80.5 & 69.3 & 19.9 & 56.6 & 6.7 & 3.3 & 40.0 & 54.0 & 17.3 & 21.9 & 23.9 & 72.2 & 42.5 & 48.8 \\
Qwen2.5-Math-7B\tablesource{yang2024qwen25math} & 61.0 & 57.9 & 16.2 & 45.0 & 10.0 & 16.7 & 42.5 & 64.2 & 15.4 & 28.0 & 29.5 & 73.5 & 39.8 & 47.0 \\
\midrule
\rowcolor{tablegroupgray}\multicolumn{15}{c}{\textbf{Zero-Style Reasoners: Curated Coding Data}} \\
\midrule
AceCoder-RM\tablesource{zeng2025acecoder} & 78.0 & 66.4 & 27.5 & 57.3 & 13.3 & 3.3 & 27.5 & 62.6 & 29.4 & 29.0 & 27.5 & 83.7 & 43.2 & 52.9 \\
AceCoder-Rule\tablesource{zeng2025acecoder} & 80.5 & \underline{70.4} & 29.0 & 60.0 & 6.7 & 6.7 & 40.0 & 62.8 & 27.6 & 27.4 & 28.5 & 74.9 & 45.7 & 52.3 \\
CodeR1-12k\tablesource{liuzhang2025coder1} & 81.1 & \textbf{73.5} & 29.3 & 61.3 & 13.3 & 3.3 & 37.5 & 74.0 & 35.7 & 36.9 & 33.5 & 76.6 & 45.4 & 54.2 \\
\midrule
\rowcolor{tablegroupgray}\multicolumn{15}{c}{\textbf{Zero-Style Reasoners: Curated Math Data}} \\
\midrule
PRIME-Zero\tablesource{cui2025process} & 49.4 & 51.1 & 11.0 & 37.2 & \textbf{23.3} & \textbf{23.3} & \textbf{67.5} & 81.2 & 37.9 & \underline{41.8} & \textbf{45.8} & 83.6 & 47.3 & 53.5 \\
SimpleRL-Zoo\tablesource{zeng2025simplerl} & 73.2 & 63.2 & 25.6 & 54.0 & 16.7 & 3.3 & 57.5 & 77.0 & 35.7 & 41.0 & 38.5 & \underline{91.7} & 44.6 & 57.2 \\
ORZ\tablesource{hu2025open} & 80.5 & 64.3 & 22.0 & 55.6 & 13.3 & 16.7 & 60.0 & \underline{81.8} & 32.7 & \textbf{45.0} & 41.6 & \textbf{92.1} & 42.1 & \underline{57.9} \\
\midrule
\rowcolor{tablegroupgray}\multicolumn{15}{c}{\textbf{Absolute Zero Training: w/ No Curated Data}} \\
\midrule
AZR-Base\tablesource{zhao2025absolutezeroreinforcedselfplay} & 71.3 & 69.1 & 25.3 & 55.2 & 13.3 & 13.3 & 52.5 & 74.4 & 38.2 & 38.5 & 38.4 & 84.5 & \underline{49.5} & 56.9 \\
AZR-Coder\tablesource{zhao2025absolutezeroreinforcedselfplay} & \textbf{83.5} & 69.6 & \underline{31.7} & \underline{61.6} & 20.0 & 10.0 & 57.5 & 72.6 & 36.4 & 38.2 & 39.1 & 72.5 & 44.6 & 54.4 \\
CPM\"obius-Math\tablesource{li2026cpmobius} & 79.6 & 63.4 & 21.1 & 54.7 & 11.8 & 9.6 & 55.6 & \textbf{84.2} & \textbf{44.9} & 38.3 & 40.7 & 85.4 & 47.6 & 57.1 \\
\midrule
\rowcolor{moreblue}\multicolumn{15}{c}{\textbf{\more{}: Self-Evolution w/ No Curated Data (Ours)}} \\
\midrule
\textsc{MORE}-7B & 76.7 & 66.8 & 30.8 & 58.1 & \underline{21.0} & 11.0 & 58.5 & 79.2 & 43.6 & 40.4 & 42.3 & 88.6 & \textbf{51.5} & \textbf{60.1} \\
\textsc{MORE}-Coder-7B & \underline{81.7} & 68.8 & \textbf{38.0} & \textbf{62.8} & 17.7 & \underline{17.7} & \underline{61.0} & 78.2 & \underline{44.5} & 37.5 & \underline{42.8} & 77.8 & 46.0 & 57.3 \\
\bottomrule
\end{tabular}
}
\end{table*}

Table~\ref{tab:main-results} compares models under their reported training regimes. Each row reports the three code benchmarks, six mathematical reasoning benchmarks, and supplemental GSM8K and MMLU-Pro results. \more{} performs Stage-2 self-evolution without curated data. Across complementary Base and Coder initializations, both variants improve on their corresponding Qwen2.5-7B backbones in CAvg, MAvg, GSM8K, MMLU-Pro, and AVG. \moremodel{}-7B attains the highest overall AVG (60.1) and the best MMLU-Pro score (51.5), while also reaching the second-best AIME 2024 result. It exceeds AZR-Base on every aggregate metric in Table~\ref{tab:main-results}.

Notably, \moremodel{}-Coder-7B sets the strongest aggregate coding results in Table~\ref{tab:main-results}, leading both LiveCodeBench v1--v5 and CAvg. Its capability gains extend across mathematical reasoning, where it is second-best on AIME 2025, AMC 2023, Minerva Math, and MAvg. It also leads AZR-Coder on LiveCodeBench, CAvg, MAvg, GSM8K, MMLU-Pro, and AVG. These results span both Base and Coder backbones: MORE-7B achieves the highest overall AVG score, while MORE-Coder-7B achieves the highest mean score of Coding benchmarks.

\begin{figure*}[t]
\centering
\begin{minipage}[b]{0.45\textwidth}
  \centering
  \includegraphics[width=\linewidth]{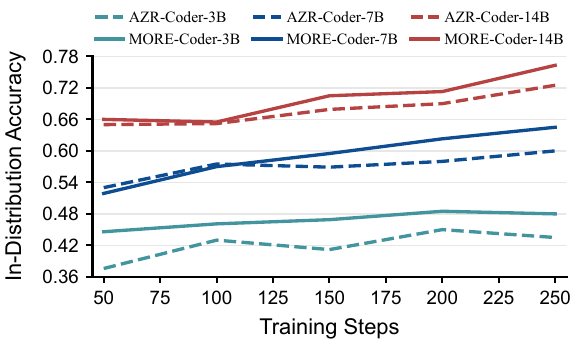}
  \par\vspace{-0.2em}\textbf{(a)}
\end{minipage}\hspace{0.015\textwidth}%
\begin{minipage}[b]{0.535\textwidth}
  \centering
  \footnotesize
  \setlength{\tabcolsep}{1.7pt}
  \renewcommand{\arraystretch}{1.30}
  \resizebox{\linewidth}{!}{%
  \begin{tabular}{@{}ll|
    >{\small}c
    >{\small}c
    >{\small}c
    >{\small}c
    >{\columncolor{black!8}\small}c@{}}
  \toprule
  \textbf{Model} & \textbf{Variant}
  & \textbf{CAvg} & \textbf{MAvg} & \textbf{GSM8K}
  & \textbf{M-Pro} & \textbf{AVG} \\
  \midrule

  Qwen2.5-Coder-3B & &
  51.0 & 18.8 & 69.4 & 38.3 & 44.4 \\

  AZR-Coder-3B & +AZR &
  54.9 & 26.5 & 71.3 & 40.3 & 48.3 \\

  \textsc{MORE}-Coder-3B & +MORE &
  55.6\moregain{+0.7} &
  28.0\moregain{+1.5} &
  73.3\moregain{+2.0} &
  42.5\moregain{+2.2} &
  49.9\moregain{+1.6} \\

  Qwen2.5-14B-Base & &
  55.5 & 28.7 & 84.0 & 40.1 & 52.1 \\

  AZR-14B & +AZR &
  58.2 & 41.9 & 85.2 & 41.9 & 56.8 \\

  \textsc{MORE}-14B & +MORE &
  60.4\moregain{+2.2} &
  42.2\moregain{+0.3} &
  87.6\moregain{+2.4} &
  42.3\moregain{+0.4} &
  58.1\moregain{+1.3} \\

  Qwen2.5-Coder-14B & &
  60.0 & 20.3 & 81.7 & 42.8 & 51.2 \\

  AZR-Coder-14B & +AZR &
  63.6 & 43.0 & 83.5 & 44.7 & 58.7 \\

  \textsc{MORE}-Coder-14B & +MORE &
  66.7\moregain{+3.1} &
  46.0\moregain{+3.0} &
  85.1\moregain{+1.6} &
  45.8\moregain{+1.1} &
  60.9\moregain{+2.2} \\

  \bottomrule
  \end{tabular}%
  }
  \par\vspace{0.2em}\textbf{(b)}
\end{minipage}
\vspace{-1.5em}
\caption{In-distribution and out-of-distribution reasoning task performance. \textbf{(a)} Average scores on CruxEval-I, CruxEval-O, and LiveCodeBench-Execution, corresponding to abduction, deduction, and deduction task types, respectively, evaluate in-distribution abilities of AZR and \more{} throughout training across Coder 3B, 7B, and 14B backbones. \textbf{(b)} Aggregate out-of-distribution scores for Coder-3B, Base-14B, and Coder-14B variants. Complete results appear in Appendix~\ref{app:detailed-scaling-results}.}
\label{fig:scaling-id-ood-results}
\end{figure*}

\noindent\textbf{Cross-Scale Generalization.} Figure~\ref{fig:scaling-id-ood-results} shows that this advantage extends beyond the two 7B initializations in Table~\ref{tab:main-results}. Across Coder-3B, Base-14B, and Coder-14B, \more{} exceeds the matched AZR model on every aggregate out-of-distribution measure in Figure~\ref{fig:scaling-id-ood-results}(b). The gains cover code and mathematical reasoning as well as GSM8K and MMLU-Pro, with the largest improvements appearing for the Coder-14B variant, where \more{} raises CAvg, MAvg, and AVG by 3.1, 3.0, and 2.2 points, respectively. Across the five evaluated Base and Coder configurations from 3B to 14B, \more{} improves the reported out-of-distribution aggregates over the corresponding AZR variants.

The in-distribution trajectories in Figure~\ref{fig:scaling-id-ood-results}(a) provide a complementary training-time view. Across the Coder-3B, Coder-7B, and Coder-14B trajectories, \more{} rises more strongly and remains above its matched AZR curve through the later training course. Regardless of model scales and backbone types, \more{} yields stronger in-distribution trajectories during self-evolution and higher out-of-distribution scores on the evaluation benchmarks.

\begin{figure}[t]
\centering
\includegraphics[width=0.7\textwidth]{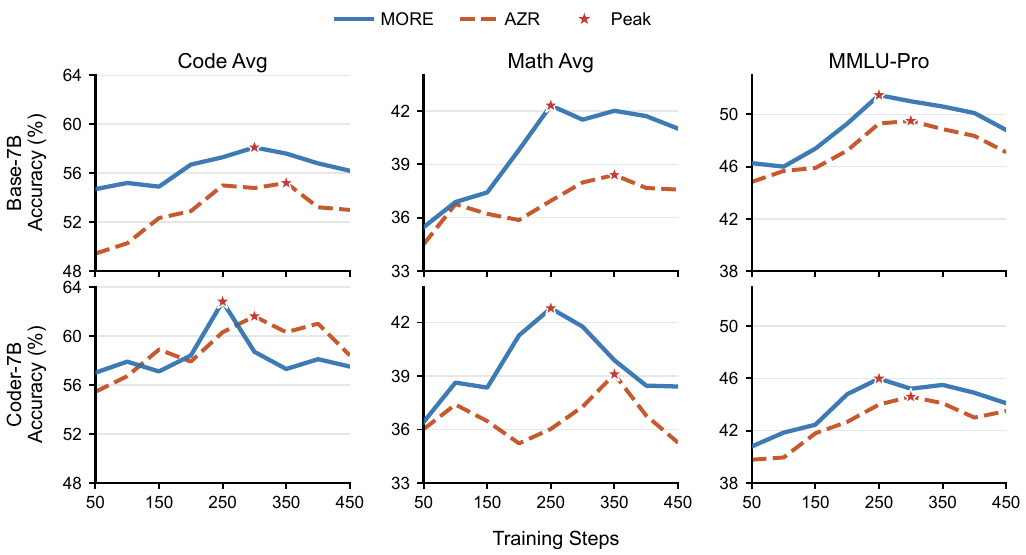}
\vspace{-1.3em}
\caption{Long-horizon out-of-distribution trajectories for matched Base-7B and Coder-7B models. Columns report Code Avg, Math Avg, and MMLU-Pro (as general reasoning).}
\label{fig:longhorizon-ood-results}
\end{figure}

\noindent\textbf{Long-Horizon OOD Dynamics.} The long-horizon comparison in Figure~\ref{fig:longhorizon-ood-results} shows that these out-of-distribution gains are sustained throughout optimization. Over the matched Base-7B and Coder-7B training trajectories, \more{} is above AZR for the overwhelming majority of evaluated checkpoints across coding, mathematical, and general reasoning. In particular, \more{} exceeds AZR at every measured MMLU-Pro checkpoint for both backbone families. The curves also reveal earlier capability formation: in nearly every paired metric trajectory, \more{} reaches its strongest observed out-of-distribution performance at least one full 50-step evaluation interval before AZR. These complementary patterns show that the improvement is not only a favorable endpoint, but also a broadly stronger and earlier-emerging reasoning trajectory.

\subsection{Training and Inference Efficiency Analysis}

\begin{figure*}[t]
\centering
\includegraphics[width=\textwidth]{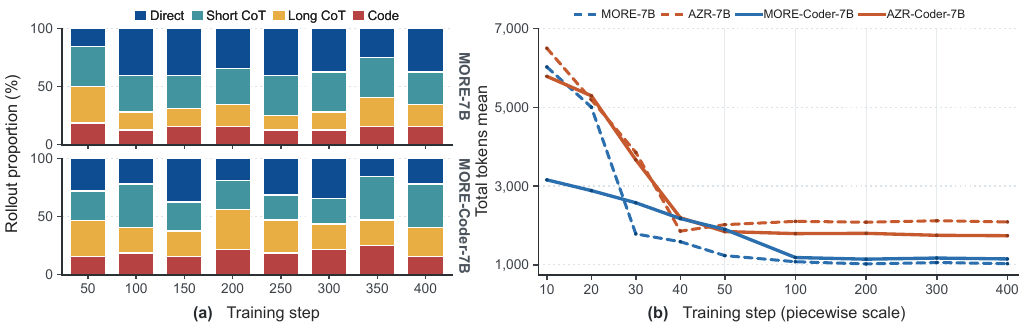}
\vspace{-0.6em}
\caption{Training-time metrics. \textbf{(a)} Rollouts' mode proportions for \more{} 7B models. \textbf{(b)} Mean prompt-plus-response tokens per rollout for matched \more{} and AZR runs.}
\label{fig:training-time-metrics}
\end{figure*}

\noindent\textbf{Adaptive Training Efficiency.} Figure~\ref{fig:training-time-metrics} (a) shows that autonomous mode allocation remains diverse throughout training. All four modes retain substantial probability mass across the measured checkpoints, and no single mode dominates either the Base or Coder trajectory. The mode interface remains an active policy decision, with all reasoning forms contributing throughout self-evolution.

Figure~\ref{fig:training-time-metrics} (b) further shows rapid token calibration during the first 50 updates, followed by a largely stable regime from approximately step 100 onward. In this regime, the mean prompt-plus-response token count of \more{} remains around 50\% of matched AZR for Base-7B and around 67\% for Coder-7B. Along with the earlier out-of-distribution convergence in Figure~\ref{fig:longhorizon-ood-results}, these trajectories reveal that adaptive mode selection and budget control jointly improve training efficiency: \more{} develops strong reasoning capability earlier while sustaining substantially lower per-rollout token demand. Additional training- and validation-time measurements appear in Appendix~\ref{app:training-time-efficiency}.

\begin{table*}[!t]
\centering
\caption{Out-of-distribution evaluation efficiency. \textbf{Response Length Avg} is the sample-mean number of generated response tokens, and \textbf{Wall Time} is the total elapsed time for the corresponding benchmark suite. The best results within each Base or Coder model family are highlighted in \textbf{bold}.}
\label{tab:ood-evaluation-efficiency}
\small
\setlength{\tabcolsep}{6pt}
\renewcommand{\arraystretch}{1.12}
\begin{tabular}{llrrrr}
\toprule
\textbf{Metric} & \textbf{Model} & \textbf{Code Avg} & \textbf{Math Avg} & \textbf{GSM8K} & \textbf{MMLU-Pro} \\
\midrule
\textbf{Response Length Avg} & Qwen2.5-7B-Base & 394.2 & 3975.7 & 499.2 & 553.3 \\
& AZR-7B & 603.7 & 4468.2 & 412.6 & 485.4 \\
& \moremodel{}-7B & \textbf{291.5} & \textbf{3173} & \textbf{195.9} & \textbf{351.7} \\
\cline{2-6}
& Qwen2.5-7B-Coder & 366.7 & 4225.1 & 418.5 & 501.0 \\
& AZR-Coder-7B & 537.4 & 5769.3 & 322.5 & 537.9 \\
& \moremodel{}-Coder-7B & \textbf{191.73} & \textbf{3015.9} & \textbf{105.1} & \textbf{303.4} \\
\midrule
\textbf{Wall Time} & Qwen2.5-7B-Base & 97m 45s & 54m 19s & 33m 29s & 44m 51s \\
& AZR-7B & 164m 27s & 62m 46s & 24m 40s & 38m 1s \\
& \moremodel{}-7B & \textbf{68m 23s} & \textbf{49m 27s} & \textbf{15m 6s} & \textbf{27m 22s} \\
\cline{2-6}
& Qwen2.5-7B-Coder & 94m 26s & 62m 22s & 30m 48s & 40m 53s \\
& AZR-Coder-7B & 143m 2s & 88m 48s & 23m 44s & 43m 53s \\
& \moremodel{}-Coder-7B & \textbf{54m 36s} & \textbf{51m 39s} & \textbf{7m 44s} & \textbf{24m 45s} \\
\bottomrule
\end{tabular}
\end{table*}

\noindent\textbf{Inference-Time Efficiency.} Table~\ref{tab:ood-evaluation-efficiency} extends the efficiency analysis to out-of-distribution evaluation. Across the code, mathematical, GSM8K, and MMLU-Pro groups, \more{} models produce shorter responses and complete evaluation in less wall-clock time than their matched AZR counterparts on both Base-7B and Coder-7B backbones. Averaged equally over the eight matched backbone--benchmark pairs, \more{} reduces Response Length Avg by 48.0\% and Wall Time by 45.1\% relative to AZR. These reductions accompany higher benchmark scores across the corresponding reported results in Table~\ref{tab:main-results}, demonstrating that the \more{} training framework significantly improves inference efficiency while strengthening reasoning performance. As a supplement, Appendix~\ref{app:research-questions} provides a more detailed investigation of test-time adaptive reasoning behavior.

\subsection{Ablation Study and Analysis}

\begin{table*}[!t]
\caption{Ablation results on Qwen2.5-Coder-7B. \textbf{+10K} increases the maximum response length to 10,240 tokens during training process. The best result in each column is shown in \textbf{bold}.}
\label{tab:ablation-results}
\centering
\small
\setlength{\tabcolsep}{1.0pt}
\renewcommand{\arraystretch}{1.04}
\resizebox{\textwidth}{!}{%
\begin{tabular}{@{}l|*{4}{c}|*{7}{c}|*{2}{c}|>{\columncolor{black!8}}c@{}}
\toprule
\textbf{Method} & \textbf{\heplus} & \textbf{\mbppplus} & \textbf{LCB} & \textbf{CAvg} & \textbf{A24} & \textbf{A25} & \textbf{AMC} & \textbf{M500} & \textbf{Min.} & \textbf{Oly.} & \textbf{MAvg} & \textbf{GSM8K} & \textbf{M-Pro} & \textbf{AVG} \\
\midrule
Qwen2.5-Coder-7B (Base) & 80.5 & 69.3 & 19.9 & 56.6 & 6.7 & 3.3 & 40.0 & 54.0 & 17.3 & 21.9 & 23.9 & 72.2 & 42.5 & 48.8 \\
Base +10K & 80.9 & \textbf{69.7} & 21.4 & 57.3 & 6.7 & 3.3 & 42.4 & 55.2 & 17.1 & 22.2 & 24.5 & 71.9 & 42.8 & 49.1 \\
Base + Stage 1 & 77.3 & 65.7 & 13.7 & 52.2 & 3.3 & 0.0 & 34.8 & 50.1 & 11.5 & 19.6 & 19.9 & 70.0 & 39.1 & 45.3 \\
Base + Stage 2 & 81.1 & 67.4 & 20.4 & 56.3 & 3.3 & 3.3 & 39.4 & 52.7 & 17.8 & 19.5 & 22.7 & 71.6 & 40.7 & 47.8 \\
AZR-Coder-7B & 83.5 & 69.6 & 31.7 & 61.6 & \textbf{20.0} & 10.0 & 57.5 & 72.6 & 36.4 & 38.2 & 39.1 & 72.5 & 44.6 & 54.4 \\
AZR +10K & \textbf{84.0} & 68.3 & 32.6 & 61.6 & \textbf{20.0} & 13.3 & 58.0 & 72.9 & 36.8 & \textbf{38.3} & 39.9 & 72.7 & 44.5 & 54.7 \\
Base + Stage 1 + AZR & 78.7 & 67.9 & 30.1 & 58.9 & 16.7 & 10.0 & 55.7 & 70.4 & 34.9 & 37.4 & 37.5 & 70.4 & 42.6 & 52.4 \\
\moremodel{}-Coder-7B & 81.7 & 68.8 & \textbf{38.0} & \textbf{62.8} & 17.7 & \textbf{17.7} & \textbf{61.0} & \textbf{78.2} & \textbf{44.5} & 37.5 & \textbf{42.8} & \textbf{77.8} & \textbf{46.0} & \textbf{57.3} \\
\bottomrule
\end{tabular}%
}
\end{table*}

Table~\ref{tab:ablation-results} shows that expanding the maximum response length from 8,096 to 10,240 tokens raises AVG by 0.3 for both the Base and AZR controls, but produces mixed movements on individual benchmarks. This limited gain suggests that simply extending the generation ceiling approaches a capacity bottleneck at the 7B scale. Also, Stage 1 is not a capability-only warm-up: its mode SFT deliberately reshapes the backbone distribution to fit four tagged response contracts. Immediately after this reshaping, CAvg falls from 56.6 to 52.2 and MAvg from 23.9 to 19.9, reducing AVG from 48.8 to 45.3. Learning the prescribed mode syntax and response forms therefore trades part of the backbone's original unconditioned coding and mathematical behavior for a stable, controllable mode prior. The same shift is visible when this prior precedes AZR, whose AVG changes from 54.4 to 52.4. Conversely, Stage 2 without Stage 1 reaches 47.8 AVG: it retains substantial reasoning ability, but has not learned the four-format prior needed for autonomous mode allocation. MATE-PPO aligns the Stage-1 prior with verifier-grounded self-evolution. The complete two-stage configuration combines stable mode-contract learning with adaptive optimization and attains the highest AVG (57.3), CAvg, MAvg, GSM8K, and MMLU-Pro.

\section{Limitations}

Our study does not include fine-grained ablations of MATE-PPO's internal components, since isolating individual terms while preserving its coupled mode--difficulty--budget dynamics did not yield a meaningful controlled comparison. \more{} uses four predefined response contracts and online difficulty estimates from task and rollout statistics. Its efficiency metrics are too coarse to resolve relative changes among finer-grained time components across training and evaluation.

\section{Conclusion}

We introduced \more, a two-stage self-evolving training framework that learns reasoning mode and computational effort as coupled policy decisions. Stage 1 establishes compact response contracts for direct answering, short and long chain-of-thought, and code-based reasoning. Stage 2 self-evolves this prior on generated, verifiable task streams without curated data, with MATE-PPO optimizing mode selection, response generation, and effort calibration within one trajectory objective. Across coding, mathematical, and general-reasoning benchmarks, \more{} achieves strong and consistent performance on Base and Coder backbones and model scales from 3B to 14B. \more{} reaches strong capability earlier during self-evolution while sustaining lower rollout token demand, shorter evaluation responses, and lower inference latency. Taken together, these research results establish \more{} as a framework for task-adaptive reasoning form and effort, advancing efficient self-evolving training and its models' reasoning through joint mode--budget learning.
\section*{AI Use Statement}
\label{app:ai-use-statement}
Generative AI models were used to generate the code-reasoning and long-CoT rationales in the Stage-1 SFT dataset; construction details are provided in Appendix~\ref{app:sft-datasets}. At the same time, we also utilized generative AI to assist in refining the wording of parts of the paper and in writing some of the complex code. We did \textbf{NOT} use generative AI to conduct fully auto-research (including the writing of the paper).
\begingroup
\makeatletter
\setlength{\@fptop}{0pt}
\makeatother
\endgroup
\bibliography{frsa_references}
\bibliographystyle{iclr2027_conference}

\newpage
\appendix
\makeatletter
\setlength{\@fptop}{0pt}
\makeatother
\section*{Appendix}

\begin{center}
\large{\textcolor[HTML]{C9A5DE}{\textbf{Appendix Contents}}}
\end{center}
\vspace{0.8em}
\begingroup
\normalsize
\appendixtocsection{app:training-metrics}{A. Training Hyperparameters and Metric Definitions}
\appendixtocsubsection{app:training-hyperparameters}{A.1 Training Hyperparameters}
\appendixtocsubsection{app:backbone-provenance}{A.2 Backbone Provenance}
\appendixtocsubsection{app:metric-definitions}{A.3 Metric Definitions}
\appendixtocsection{app:sft-datasets}{B. SFT Datasets in Stage 1}
\appendixtocsection{app:stage2-training}{C. Stage 2 Training Process}
\appendixtocsubsection{app:six-rollout-streams}{C.1 Six Proposer--Solver Rollout Streams}
\appendixtocsubsection{app:execution-feedback}{C.2 Execution-Grounded Feedback}
\appendixtocsubsection{app:mate-algorithm}{C.3 MATE-PPO}
\appendixtocsection{app:runtime-mapping}{D. Mode-Token Interface and MATE-PPO Runtime}
\appendixtocsection{app:rollout-clustering}{E. Rollout Clustering and Difficulty Assignment}
\appendixtocsection{app:reward-defaults}{F. Reward Defaults and Runtime Settings}
\appendixtocsection{app:evaluation-rules}{G. Evaluation Protocol and Reporting Rules}
\appendixtocsubsection{app:longhorizon-checkpoint-selection}{G.1 Long-Horizon Checkpoint Selection}
\appendixtocsubsection{app:controlled-ablation-protocol}{G.2 Controlled Ablation Protocol}
\appendixtocsubsection{app:seed-averaged-main-results}{G.3 Seed-Averaged Main Results}
\appendixtocsection{app:detailed-scaling-results}{H. Scaling Results Across Model Sizes}
\appendixtocsubsection{app:scaling-ood-results}{H.1 Out-of-Distribution Results}
\appendixtocsubsection{app:in-distribution-results}{H.2 In-Distribution Results}
\appendixtocsubsection{app:training-time-efficiency}{H.3 Training-Time Efficiency and Difficulty Dynamics}
\appendixtocsection{app:research-questions}{I. Research Questions}
\appendixtocsubsection{app:test-time-behavior}{I.1 Test-Time Adaptive Reasoning Behavior}
\appendixtocsubsection{app:test-time-efficiency}{I.2 Test-Time Response-Efficiency Patterns}
\appendixtocsubsection{app:terminal-return-dynamics}{I.3 Interpretable Terminal-Return Dynamics}
\appendixtocsubsection{app:task-stream-diversity}{I.4 Self-Generated Task-Stream Diversity}
\appendixtocsubsection{app:mode-budget-calibration}{I.5 Mode-Conditioned Budget Calibration}
\appendixtocsubsection{app:stream-mode-utility}{I.6 Role-Sensitive Mode Utility}
\endgroup
\clearpage

\section{Training Hyperparameters and Metric Definitions}
\label{app:training-metrics}

\subsection{Training Hyperparameters}
\label{app:training-hyperparameters}
Table~\ref{tab:training-hyperparameters} reports the configuration for the main experiments. The two backbones use the same Stage-1 and Stage-2 profile; unless a result is explicitly identified as an ablation, these settings remain fixed.

\begin{table}[H]
\caption{Training hyperparameters for the reported main runs. The six Stage-2 streams comprise proposer and solver rollouts for each executable task family.}
\label{tab:training-hyperparameters}
\centering
\footnotesize
\setlength{\tabcolsep}{4pt}
\begin{tabular}{p{0.54\linewidth}p{0.37\linewidth}}
\toprule
\textbf{Parameter} & \textbf{Value} \\
\midrule
\multicolumn{2}{l}{\textit{Shared runtime}} \\
Accelerators & $4\times$ NVIDIA A100 (80 GB) \\
Numerical precision & BF16 mixed precision \\
Stage-2 rollout tensor parallelism & TP $= 2$ \\
\addlinespace
\multicolumn{2}{l}{\textit{Stage 1: mode-format SFT}} \\
SFT method & All-target LoRA \\
Sequence cutoff & 4,096 tokens \\
Learning rate / schedule & $2\times10^{-4}$ / cosine, 0.1 warm-up ratio \\
Per-device batch / gradient accumulation & 4 / 2 \\
Effective batch / training epochs & 64 / 6 \\
Validation split / optimizer steps & 10\% / 3,654 \\
\addlinespace
\multicolumn{2}{l}{\textit{Stage 2: MATE-PPO and self-evolution}} \\
Maximum prompt / response length & 8,096 / 8,096 tokens \\
Budget-controller scale & 6,144 tokens \\
Maximum batched tokens & 16,384 \\
Base task batch / rollout streams & 32 / 6 \\
Sampled trajectories per update & 192 \\
PPO minibatch / microbatch per GPU & 32 / 2 \\
Advantage estimator & REINFORCE++ \\
Actor learning rate / gradient clip & $10^{-6}$ / 1.0 \\
PPO clip ratio / epochs & 0.2 / 1 \\
KL loss / coefficient & disabled / 0 \\
Entropy coefficient & 0.001 \\
Rollouts per prompt / temperature / top-$p$ & 1 / 1.0 / 1.0 \\
Seed-batch factor / maximum retained programs & 4 / 16,384 \\
Samples for task-accuracy estimation & 8 \\
Checkpoint / full-benchmark cadence & every 10 / every 50 steps \\
\bottomrule
\end{tabular}
\end{table}

\begingroup
\makeatletter
\setlength{\@fptop}{0pt}
\makeatother
\begin{table*}[t]
\subsection{Backbone Provenance}
\label{app:backbone-provenance}
\caption{Training-data source and backbone provenance for the models in Table~\ref{tab:main-results} and Figure~\ref{fig:scaling-id-ood-results}. For \more{} rows, Stage-2 entries state whether self-evolution uses curated data.}
\label{tab:backbone-provenance}
\centering
\footnotesize
\setlength{\tabcolsep}{4pt}
\renewcommand{\arraystretch}{1.10}
\begin{tabular}{@{}>{\raggedright\arraybackslash}p{0.30\textwidth}>{\raggedright\arraybackslash}p{0.43\textwidth}>{\raggedright\arraybackslash}p{0.21\textwidth}@{}}
\toprule
\textbf{Model} & \textbf{Data Curation} & \textbf{Base Model} \\
\midrule
Qwen2.5-7B-Base\tablesource{qwen2025qwen25technicalreport} & N/A (off-the-shelf base checkpoint) & Qwen2.5-7B-Base \\
Qwen2.5-Coder-7B\tablesource{hui2024qwen25coder} & N/A (off-the-shelf base checkpoint) & Qwen2.5-Coder-7B \\
Qwen2.5-Math-7B\tablesource{yang2024qwen25math} & N/A (off-the-shelf base checkpoint) & Qwen2.5-Math-7B \\
Qwen2.5-Coder-3B\tablesource{hui2024qwen25coder} & N/A (off-the-shelf base checkpoint) & Qwen2.5-Coder-3B \\
Qwen2.5-14B-Base\tablesource{qwen2025qwen25technicalreport} & N/A (off-the-shelf base checkpoint) & Qwen2.5-14B-Base \\
Qwen2.5-Coder-14B\tablesource{hui2024qwen25coder} & N/A (off-the-shelf base checkpoint) & Qwen2.5-Coder-14B \\
AceCoder-RM\tablesource{zeng2025acecoder} & 22K code examples & Qwen2.5-7B-Coder \\
AceCoder-Rule\tablesource{zeng2025acecoder} & 22K code examples & Qwen2.5-7B-Coder \\
CodeR1-12k\tablesource{liuzhang2025coder1} & 2K LeetCode + 10K TACO pairs & Qwen2.5-7B-Instruct-1M \\
PRIME-Zero\tablesource{cui2025process} & 457K math + 27K code problems & Qwen2.5-Math-7B-Base \\
SimpleRL-Zoo\tablesource{zeng2025simplerl} & 8.5K MATH pairs \citep{hendrycks2021math} & Qwen2.5-7B-Base \\
ORZ\tablesource{hu2025open} & 57K STEM + math samples & Qwen2.5-7B-Base \\
AZR-Base\tablesource{zhao2025absolutezeroreinforcedselfplay} & No curated data & Qwen2.5-7B-Base \\
AZR-Coder\tablesource{zhao2025absolutezeroreinforcedselfplay} & No curated data & Qwen2.5-7B-Coder \\
AZR-Coder-3B\tablesource{zhao2025absolutezeroreinforcedselfplay} & No curated data & Qwen2.5-Coder-3B \\
AZR-14B\tablesource{zhao2025absolutezeroreinforcedselfplay} & No curated data & Qwen2.5-14B-Base \\
AZR-Coder-14B\tablesource{zhao2025absolutezeroreinforcedselfplay} & No curated data & Qwen2.5-Coder-14B \\
CPM\"obius-Math\tablesource{li2026cpmobius} & No external data during co-evolution; 4K PRIME data for Coach warm-up & Qwen2.5-Math-7B-Instruct \\
\moremodel{}-7B & Stage 1: 10.8K AQuA-RAT mode-format SFT; Stage 2: self-evolution without curated data & Qwen2.5-7B-Base \\
\moremodel{}-Coder-7B & Stage 1: 10.8K AQuA-RAT mode-format SFT; Stage 2: self-evolution without curated data & Qwen2.5-Coder-7B \\
\moremodel{}-Coder-3B & Stage 1: 10.8K AQuA-RAT mode-format SFT; Stage 2: self-evolution without curated data & Qwen2.5-Coder-3B \\
\moremodel{}-14B & Stage 1: 10.8K AQuA-RAT mode-format SFT; Stage 2: self-evolution without curated data & Qwen2.5-14B-Base \\
\moremodel{}-Coder-14B & Stage 1: 10.8K AQuA-RAT mode-format SFT; Stage 2: self-evolution without curated data & Qwen2.5-Coder-14B \\
\bottomrule
\end{tabular}
\end{table*}
\clearpage
\endgroup

\subsection{Metric Definitions}
\label{app:metric-definitions}
\paragraph{Capability aggregates.}
Coding evaluation uses 164 \humanevalplus{} and 378 \mbppplus{} problems from EvalPlus together with the cumulative LiveCodeBench v1--v5 views containing 400, 511, 612, 713, and 880 problems. EvalPlus uses greedy generation. LiveCodeBench uses one-sample pass@1 with temperature 0, top-$p=1$, an 8,096-token generation limit, and a 60-second execution timeout. Let $h^+$, $m^+$, and $\ell_{\mathrm{avg}(v1\text{--}v5)}$ denote the \humanevalplus{}, \mbppplus{}, and mean LiveCodeBench v1--v5 pass@1 scores in percentage points. Then
\begin{equation}
  \mathrm{Code\ Avg}=\frac{h^+ + m^+ + \ell_{\mathrm{avg}(v1\text{--}v5)}}{3}.
  \label{eq:code-avg}
\end{equation}

Mathematical evaluation uses AIME 2024 (30 problems), AIME 2025 (30), AMC 2023 (40), MATH-500 (500), OlympiadBench (675), and Minerva Math (272). Decoding uses one sample with temperature 0, top-$p=0.95$, a 16,000-token limit, and seed 2. For the six single-sample pass@1 scores $a_1,\ldots,a_6$ measured in percentage points,
\begin{equation}
  \mathrm{Math\ Avg}=\frac{1}{6}\sum_{j=1}^{6}a_j.
  \label{eq:math-avg}
\end{equation}

GSM8K uses the official \texttt{main} test split and MMLU-Pro uses its official test split. Both use one greedy sample per task.

\paragraph{Out-of-distribution evaluation efficiency.}
For a benchmark suite with generated responses $y_1,\ldots,y_N$, the response-length average in Table~\ref{tab:ood-evaluation-efficiency} is
\begin{equation}
  \mathrm{Response\ Length\ Avg}=\frac{1}{N}\sum_{i=1}^{N}|y_i|,
  \label{eq:response-length-avg}
\end{equation}
where $|y_i|$ is the number of generated response tokens. Wall time is the total elapsed evaluation time for the corresponding suite.

\section{SFT Datasets in Stage 1}
\label{app:sft-datasets}

\begin{figure*}[t]
  \centering
  \includegraphics[width=\textwidth]{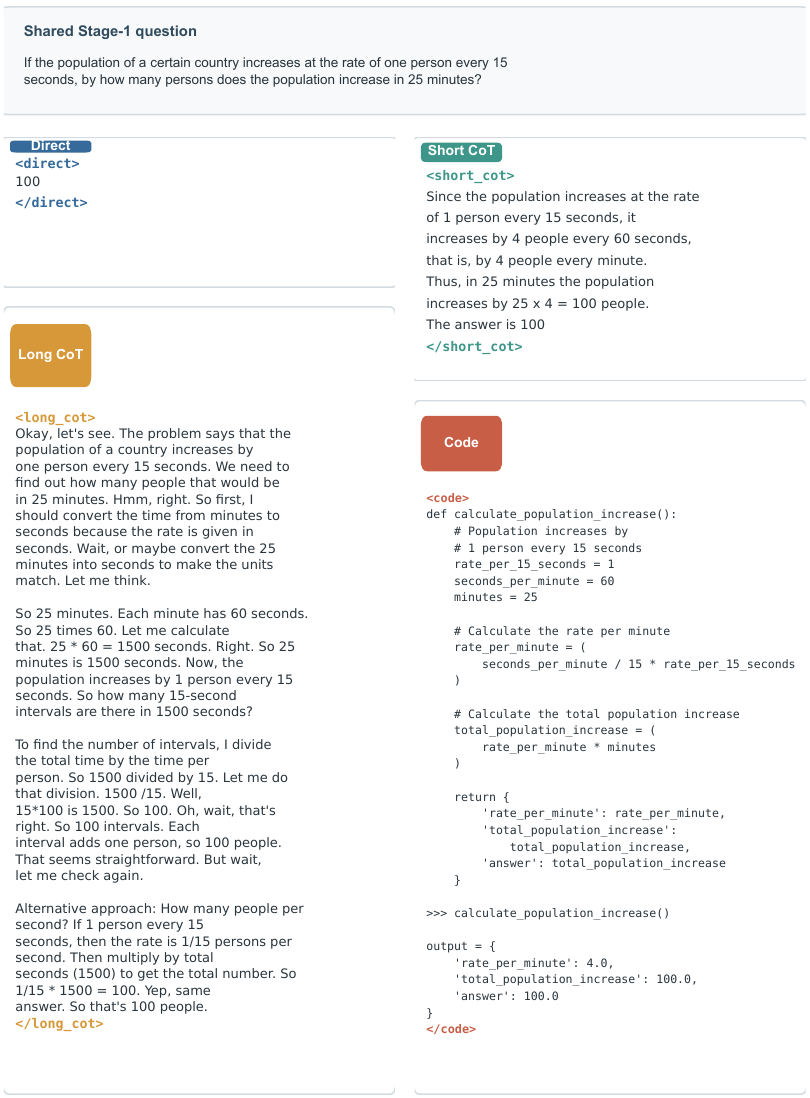}
  \caption{A shared Stage-1 question expressed through the four response contracts. Direct, Short CoT, Long CoT, and Code share the same target answer while instantiating distinct reasoning forms.}
  \label{fig:stage1-mode-contract-example}
\end{figure*}

The Stage-1 SFT corpus starts from 10,826 AQuA-RAT source questions: 3,000 multiple-choice and 7,826 open-form questions \citep{ling2017programinductionrationalegeneration}. Following the four-mode construction procedure of ARM \citep{wu2025armadaptivereasoningmodel}, every source question is rendered once under each of the \texttt{direct}, \texttt{short\_cot}, \texttt{long\_cot}, and \texttt{code} response contracts. This produces 10,826 supervised records per contract and 43,304 SFT records in total (12,000 multiple-choice and 31,304 open-form records). Code reasoning rationales are generated with GPT-5, while long-CoT rationales are generated with DeepSeek-R1 \citep{deepseekai2026deepseekr1incentivizingreasoningcapability}. The corpus uses a 4,096-token cutoff. The mean token counts are 12.7 for Direct, 110.4 for Short CoT, 252.8 for Code, and 1,226.5 for Long CoT. Figure~\ref{fig:stage1-mode-contract-example} illustrates the four targets for one shared question.

\clearpage

\section{Stage 2 Training Process}
\label{app:stage2-training}

Stage 2 uses one policy in a closed proposer--solver loop. At every update, the policy first constructs executable tasks, then solves tasks drawn from the corresponding typed pools. Both roles generate a mode token followed by a response, receive execution-grounded terminal feedback, and contribute trajectories to the same policy update. This loop continuously refreshes the training distribution as the policy improves.

\subsection{Six Proposer--Solver Rollout Streams}
\label{app:six-rollout-streams}

The task space combines three verifiable reasoning families with two policy roles, yielding six rollout streams. Let $G_\tau$ and $S_\tau$ denote proposer and solver rollouts for family $\tau\in\{\mathrm{in},\mathrm{out},\mathrm{fun}\}$. The resulting batch is
\begin{equation}
  \mathcal B_t=\mathcal B_t^{G_{\mathrm{in}}}\cup\mathcal B_t^{S_{\mathrm{in}}}
  \cup\mathcal B_t^{G_{\mathrm{out}}}\cup\mathcal B_t^{S_{\mathrm{out}}}
  \cup\mathcal B_t^{G_{\mathrm{fun}}}\cup\mathcal B_t^{S_{\mathrm{fun}}}.
  \label{eq:six-rollout-streams}
\end{equation}
The three families and their executable contracts are summarized in Table~\ref{tab:stage2-task-families}.

\begin{table}[H]
\caption{Three executable task families, each instantiated as a proposer and a solver stream.}
\label{tab:stage2-task-families}
\centering
\footnotesize
\setlength{\tabcolsep}{3pt}
\begin{tabular}{>{\raggedright\arraybackslash}p{0.19\linewidth}>{\raggedright\arraybackslash}p{0.25\linewidth}>{\raggedright\arraybackslash}p{0.22\linewidth}>{\raggedright\arraybackslash}p{0.25\linewidth}}
\toprule
\textbf{Family} & \textbf{Proposed task state} & \textbf{Solver target} & \textbf{Execution check} \\
\midrule
Input recovery (abduction) & Executable function $f$ and an input $x$; execution supplies $y=f(x)$ & Recover $\hat x$ from $(f,y)$ & Check $f(\hat x)=y$ \\
Output prediction (deduction) & Executable function $f$ paired with input $x$ & Predict $\hat y$ from $(f,x)$ & Compare $\hat y$ with $f(x)$ \\
Function induction & Inputs for a latent executable function; execution supplies visible and hidden I/O pairs & Synthesize $\hat f$ from visible pairs & Run $\hat f$ on every hidden pair \\
\bottomrule
\end{tabular}
\end{table}

For input recovery and output prediction, the proposer generates a parseable Python function and input state; the executor checks validity and materializes the corresponding output. For function induction, it proposes an input set for a latent executable function, from which visible and held-out I/O pairs are constructed. Accepted tasks enter family-specific pools, and solver batches are built from those evolving pools. Thus, new task construction and policy improvement remain coupled without requiring externally labeled Stage-2 solutions.

\subsection{Execution-Grounded Feedback}
\label{app:execution-feedback}

The proposer is rewarded for producing valid tasks that are solvable but not already trivial for the current policy. For a valid candidate $z_j$, the solver generates $K=8$ independent attempts. With family-specific executor $V_\tau$,
\begin{equation}
  \hat p_j=\frac{1}{K}\sum_{k=1}^{K}V_\tau\!\left(z_j,\tilde y_j^{(k)}\right),
  \qquad
  R_{\mathrm{prop}}(z_j)=g_j\,\mathbb I[\hat p_j>0](1-\hat p_j),
  \label{eq:proposer-learnability}
\end{equation}
where $g_j\in\{0,1\}$ is the parse, determinism, and execution-validity gate. Invalid candidates and candidates for which all sampled solutions fail receive zero proposer reward; among solvable tasks, lower current success yields larger learning credit. Every valid proposal is retained in its typed pool for subsequent solver sampling.

Solver feedback is direct execution accuracy. A format or execution failure receives zero. Otherwise,
\begin{equation}
  R_{\mathrm{acc}}=V_\tau(z,\hat y),\quad
  V_{\mathrm{in}}=\mathbb I[f(\hat x)=y],\quad
  V_{\mathrm{out}}=\mathbb I[\hat y=f(x)],\quad
  V_{\mathrm{fun}}=\prod_{(x_h,y_h)\in\mathcal H}\mathbb I[\hat f(x_h)=y_h],
  \label{eq:solver-execution-reward}
\end{equation}
where $\mathcal H$ is the held-out I/O set for function induction. The binary hidden-test rule gives function synthesis full accuracy credit only when all held-out cases pass. These verifier outcomes are returned to the same trajectories that produced the proposals or solutions and also update the online difficulty and budget state.

\subsection{MATE-PPO}
\label{app:mate-algorithm}

\paragraph{PPO base for joint mode--budget control.}
MATE-PPO uses a PPO-style clipped actor update as its optimization base and augments it with a controller-conditioned terminal return. Unlike Group Relative Policy Optimization (GRPO) \citep{shao2024deepseekmath}, which forms advantages by ranking multiple completions of a common prompt within a local group, \more{} must optimize heterogeneous proposer and solver behaviors under one shared policy. Proposer samples construct different executable tasks, whereas several solver modes can be correct and compliant with their own $(d,m)$ budget bands on the same task. Group-relative normalization would make such valid task constructions or mode contracts compete within a sampled group and remove return differences across task groups from the advantage estimator. The trajectory-level PPO objective instead preserves direct, shared-scale credit for task construction, mode selection, solution accuracy, and effort calibration across the six-stream update. It also maintains broad task coverage for the online controller without requiring repeated long rollouts from every frozen controller state.

MATE-PPO constructs a unified terminal return by integrating role-specific execution feedback with difficulty-conditioned budget calibration, mode-fit utility, and mode--budget interaction. For rollout $i$, let the role-specific execution gate be
\begin{equation}
  c_i=
  \begin{cases}
    R_{\mathrm{acc},i}=V_\tau(z_i,\hat y_i), & i\text{ is a solver rollout},\\
    g_i\,\mathbb I[\hat p_i>0], & i\text{ is a proposer rollout},
  \end{cases}
  \label{eq:role-specific-gate}
\end{equation}
where the proposer gate requires a parseable, deterministic, executable proposal with at least one successful solver attempt. The terminal return is then
\begin{equation}
  \begin{aligned}
    R_{\mathrm{base},i}&=
    \begin{cases}
      R_{\mathrm{prop},i}, & i\text{ is a proposer rollout},\\
      R_{\mathrm{acc},i}, & i\text{ is a solver rollout},
    \end{cases}\\
    R_i^{\mathrm{terminal}}&=R_{\mathrm{base},i}
    +\alpha R_{\mathrm{BD},i}+\beta R_{\mathrm{MF},i}+\gamma R_{\mathrm{MB},i}.
  \end{aligned}
  \label{eq:stage2-role-return}
\end{equation}
For solver trajectories, Equation~\ref{eq:stage2-role-return} is exactly the \more{} return in Equation~\ref{eq:more-return}. For both roles, the mode-fit and joint terms use the execution gate: $R_{\mathrm{MF},i}=c_iU(d_i,m_i)$ and $R_{\mathrm{MB},i}=c_i\operatorname{clip}[-\eta(d_i,m_i)(o_i+u_i),-0.25,0]$. The budget-deviation term retains its overlength penalty for every rollout and uses $w_u(c_i)$ for its lower-band weight. The additive mode--budget delta is written to the final valid response position, so the role-specific execution reward is counted once. The mode token remains likelihood-bearing and is optimized jointly with the response tokens.

\paragraph{Advantage estimation and mode-token credit.}
After generation, the runtime recomputes old-policy log probabilities, evaluates the six stream-specific terminal returns, and concatenates the resulting rollouts. For rollout $i$ with $T_i$ valid generated tokens, the token-level reward is
\[
  r_{i,t}=\mathbb{I}[t=T_i]R_i^{\mathrm{terminal}},
  \qquad
  G_{i,t}=\sum_{k=t}^{T_i}r_{i,k},
\]
with discount factor $\gamma=1$. Let $\mathcal V=\{(i,t):1\leq t\leq T_i\}$ be the valid-token set of the concatenated six-stream batch, and let $\mu_{\mathcal V}$ and $\sigma_{\mathcal V}$ be the mean and unbiased standard deviation of $G_{i,t}$ over this set. The REINFORCE++ advantage is
\[
  \hat A_{i,t}^{\mathrm{MORE}}=
  \frac{G_{i,t}-\mu_{\mathcal V}}
       {\sqrt{\sigma_{\mathcal V}^{2}+10^{-8}}},
  \qquad (i,t)\in\mathcal V.
\]
Padding positions are masked from both the statistics and policy loss. Thus, the six streams share one valid-token normalization set. All terminal components, including negative budget-deviation and mode--budget terms, enter $R_i^{\mathrm{terminal}}$ before return computation and normalization. The generated \texttt{<mode>} span occupies the first valid positions of the rollout and is included in the policy-loss mask. Because the return is terminal and $\gamma=1$, its tokens receive the same standardized trajectory advantage as the subsequent response tokens; the selected mode consequently receives direct policy credit through its effect on the terminal return. With $r_{i,t}(\theta)=\pi_\theta(a_{i,t}\mid s_{i,t})/\pi_{\mathrm{old}}(a_{i,t}\mid s_{i,t})$, the actor update is
\begin{equation}
  \mathcal L_{\mathrm{MATE\text{-}PPO}}(\theta)=
  \mathbb E_{i,t}\!\left[\min\!\left(r_{i,t}(\theta)\hat A_{i,t}^{\mathrm{MORE}},
  \operatorname{clip}(r_{i,t}(\theta),1-\epsilon,1+\epsilon)\hat A_{i,t}^{\mathrm{MORE}}\right)\right].
  \label{eq:appendix-mate-ppo}
\end{equation}
One optimizer step therefore improves task construction, task solving, autonomous mode choice, and effort calibration from the same self-evolving batch.

\begin{algorithm*}[t]
\SetKwComment{tcp}{\(\triangleright\) }{}
\SetAlgoShortEnd
\SetAlCapNameFnt{\small}
\SetAlCapFnt{\small}
\SetAlFnt{\small}
\DontPrintSemicolon
\caption{\textsc{MORE} Stage 2: Self-Evolving Mode--Budget Optimization with MATE-PPO}
\label{alg:mate-algorithm}
\KwIn{Stage-1 policy $\pi_\theta$; typed task pools $\{\mathcal P_\tau\}_{\tau\in\{\mathrm{in},\mathrm{out},\mathrm{fun}\}}$; family executors $\{V_\tau\}$; mode set $\modeset$; difficulty tracker $D$; mode--budget controller $B$; mode utility $U$; solver-probe count $K=8$}
\KwOut{Evolved policy $\pi_\theta$, task pools $\{\mathcal P_\tau\}$, and controller state $(D,B)$}
\While{training has not converged}{
  \textit{// --- Phase 1: Task Construction and Proposer Feedback (Secs.~\ref{app:six-rollout-streams}--\ref{app:execution-feedback}) ---}\;
  \ForEach{$\tau\in\{\mathrm{in},\mathrm{out},\mathrm{fun}\}$}{
    \ForEach{proposer rollout $i\in\mathcal B^{G_\tau}$}{
      $(d_i,B_i)\leftarrow\operatorname{Snapshot}(D,B,\tau,i)$\;
      $(m_i,z_i)\sim\pi_\theta(\cdot\mid\tau,d_i,B_i)$, where $m_i\in\modeset$ is the generated \texttt{<mode>} token\;
      $g_i\leftarrow\operatorname{Validate}_\tau(z_i)$ \tcp*{parse, determinism, and execution validity}
      \eIf{$g_i=1$}{
        $\{(\tilde m_{i,k},\tilde y_{i,k})\}_{k=1}^{K}\sim\pi_\theta(\cdot\mid z_i)$\;
        $\hat p_i\leftarrow K^{-1}\sum_{k=1}^{K}V_\tau(z_i,\tilde y_{i,k})$; \quad $R_{\mathrm{prop},i}\leftarrow\mathbb I[\hat p_i>0](1-\hat p_i)$\;
        add $z_i$ to $\mathcal P_\tau$\;
      }{$R_{\mathrm{prop},i}\leftarrow0$}\;
    }
  }
  \textit{// --- Phase 2: Solver Rollouts and Execution Feedback (Secs.~\ref{app:six-rollout-streams}--\ref{app:execution-feedback}) ---}\;
  \ForEach{$\tau\in\{\mathrm{in},\mathrm{out},\mathrm{fun}\}$}{
    sample solver tasks $\mathcal Z_\tau\subset\mathcal P_\tau$\;
    \ForEach{$z_i\in\mathcal Z_\tau$}{
      $(d_i,B_i)\leftarrow\operatorname{Snapshot}(D,B,\tau,i)$\;
      $(m_i,\hat y_i)\sim\pi_\theta(\cdot\mid z_i,d_i,B_i)$, where $m_i\in\modeset$\;
      $R_{\mathrm{acc},i}\leftarrow V_\tau(z_i,\hat y_i)$\;
    }
  }
  \textit{// --- Phase 3: MATE-PPO Return and Joint Update (Sec.~\ref{app:mate-algorithm}) ---}\;
  \ForEach{rollout $i$ in the six proposer--solver streams}{
    $c_i\leftarrow R_{\mathrm{acc},i}$ for solver rollouts, or $g_i\mathbb I[\hat p_i>0]$ for proposer rollouts\;
    compute $R_{\mathrm{BD},i}$, $R_{\mathrm{MF},i}=c_iU(d_i,m_i)$, and $R_{\mathrm{MB},i}$ using $(d_i,m_i,B_i,c_i)$\;
    $R_{\mathrm{base},i}\leftarrow R_{\mathrm{acc},i}$ for solvers or $R_{\mathrm{prop},i}$ for proposers\;
    $R_i^{\mathrm{terminal}}\leftarrow R_{\mathrm{base},i}+\alpha R_{\mathrm{BD},i}+\beta R_{\mathrm{MF},i}+\gamma R_{\mathrm{MB},i}$; attach it to the final valid token\;
  }
  $(D,B)\leftarrow\operatorname{UpdateController}(D,B,\{d_i,m_i,z_i,\hat y_i,c_i\})$ \tcp*{produces $(D_{t+1},B_{t+1})$}
  concatenate the six streams; compute valid-token $\hat A^{\mathrm{MORE}}$ by REINFORCE++ with $\gamma=1$\;
  $\theta\leftarrow\operatorname{MATE\text{-}PPO}(\theta,\hat A^{\mathrm{MORE}})$; record task, mode, budget, and timing telemetry\;
}
\end{algorithm*}
\FloatBarrier

\section{Mode-Token Interface and MATE-PPO Runtime}
\label{app:runtime-mapping}

The policy trajectory begins with a generated \texttt{<mode>} control token followed by mode-conditioned response tokens. The runtime preserves this first action together with its likelihood, task identifier, verifier score, attended response length, controller state, and budget band. The same token is consumed by reward computation and telemetry, so mode learning, response generation, and evaluation share one representation.

For each rollout, the runtime snapshots the current controller state $(D_t,B_t)$ before generation. The generated mode token selects the response contract and indexes the band in that snapshot. After verification, the terminal \more{} delta is added at the final valid response position using the same $(D_t,B_t)$ state; the controller then incorporates the outcome and correct-length observation into $(D_{t+1},B_{t+1})$ for subsequent rollouts. This composition optimizes the mode action and response tokens jointly while avoiding duplicate counting of the verifier reward. The 8,096-token rollout cap provides generation headroom, while the 6,144-token controller scale defines the normalized initialization of budget bands.

\section{Rollout Clustering and Difficulty Assignment}
\label{app:rollout-clustering}

This section specifies the full clustering path used by the longitudinal realization. It replaces a global notion of difficulty with policy-relative groups: rollouts with comparable task structure and verifier semantics share a success history, and their relative success determines the reward-routing tier. The online feature-bucket procedure continuously adapts these groups as the policy evolves.

\paragraph{Observation and task envelope.}
For rollout $i$, the reward manager constructs $o_i=(u_i,a_i,L_i,x_i,z_i)$ from a stable task identifier $u_i$, verifier correctness $a_i$, attended response length $L_i$, problem text $x_i$, and metadata $z_i$. The task role defaults to solver. Explicit family and subtype metadata take precedence. Otherwise, the runtime uses the rollout prefix to distinguish proposer from solver and the verifier suffix to assign reasoning semantics: input recovery to abduction, output prediction to deduction, and function synthesis to induction. Prompt cues complete these assignments; tasks outside the recognized semantics enter a dedicated general family/subtype that preserves them for online clustering.

\paragraph{Rollout representation.}
The representation $\phi_i$ concatenates the feature groups in Table~\ref{tab:cluster-features}. Extracted structural and semantic fields provide the task representation, while a per-UID window of the latest 200 verified observations contributes mean success, success variance, and mean response length. After each rollout is verified, its correctness and length update this state for subsequent mode and budget decisions.

\begin{table}[t]
\caption{Feature groups used by the rollout-clustering implementation.}
\label{tab:cluster-features}
\centering
\small
\begin{tabular}{p{0.20\linewidth}p{0.73\linewidth}}
\toprule
Group & Implemented coordinates \\
\midrule
Code structure & Python AST depth; branch, loop, nested-loop, call, and function counts; recursion flag; expression depth; and list/dict/set/tuple counts. Non-Python tasks are represented by the remaining structural and semantic coordinates. \\
Prompt artifacts & Code-block, test/assert, stub, I/O-example, bug, and constraint cues; word and nonempty-line counts; input, output, and pattern cues. \\
Task semantics & One-hot family and subtype; family-specific deduction-path, abduction-cause, or induction-pattern indicators; input-contract or output-state indicators; reference-text presence. \\
Verifier shape & One-hot input, output, error, and function targets together with the corresponding contract, trace, error-surface, and function-check burden indicators. \\
General/behavior & General semantic coordinates: text length, code presence, and constraint count. For every UID: rolling success mean, success variance, response-length mean, and a scalar structural score complementing the supplied coordinates. \\
\bottomrule
\end{tabular}
\end{table}

\paragraph{Family-aware online assignment.}
Clusters are partitioned first by reasoning family and verifier target. A previously observed UID always reuses its stored cluster. For a new UID, the tracker creates a new cluster in its family--verifier partition while that partition contains fewer than $K=9$ clusters. After the partition reaches $K$, the UID is assigned to the nearest centroid under unnormalized squared Euclidean distance,
\[
  g_i=\arg\min_{c\in\mathcal C(f_i,v_i)}\sum_{j\in\mathcal J}\bigl(\mu_{c,j}-\phi_{i,j}\bigr)^2,
\]
with inactive coordinates represented by zero. Once the dedicated general-family partition is full, its candidate set broadens to all existing clusters. Family and verifier target are assignment keys; role and subtype are retained as cluster metadata and enrich the cluster representation. Cluster identifiers encode family, verifier target, and a family-level creation counter. This construction keeps deduction, abduction, induction, and input/output/error/function processes distinct during normal assignment while preserving a shared assignment path for general prompts.

\paragraph{Centroid and success updates.}
Each new cluster starts with its first feature vector, size one, and success estimate equal to the first binary outcome. Thereafter, its centroid is the arithmetic running mean, $\mu_c^{(n)}=\mu_c^{(n-1)}+[\phi_i-\mu_c^{(n-1)}]/n$, and its policy-relative success is $q_c^{(n)}=0.8q_c^{(n-1)}+0.2a_i$. The difficulty score exposed to telemetry is $1-q_c$. A cluster becomes mature after eight assigned observations. Immature clusters retain the default medium tier.

\paragraph{Quantile tiering and cold-start cases.}
Tiers are refreshed after every recorded rollout over all mature clusters. With one mature cluster, that cluster is easy. With two, the higher-success cluster is easy and the lower is medium, except that an exactly zero success EMA makes the lower cluster hard. With at least three mature clusters, the implementation sorts mature-cluster success values and uses the order statistic at index $\lfloor(n-1)q\rfloor$: a cluster is hard when $q_c\leq Q_{0.33}$, easy when $q_c>Q_{0.67}$, and medium otherwise. The strict upper and non-strict lower comparisons are part of the implementation and determine tie behavior.

\paragraph{Online state evolution.}
Representations, centroids, EMA values, and tiers update as verified rollouts arrive. Stable UID assignments preserve a consistent task history, while the family--verifier partition prevents unrelated reasoning processes from sharing statistics. The resulting lightweight state can be updated throughout self-evolution and directly supplies the difficulty context used by the next mode-token decision and its budget band.

\section{Reward Defaults and Runtime Settings}
\label{app:reward-defaults}

Tables~\ref{tab:reward-defaults} and~\ref{tab:budget-bands} give the structured utility and budget initialization used by the main longitudinal run. The weights are $(\alpha,\beta,\gamma)=(0.6,1.0,0.05)$; $\lambda_d$ is $0.01$, $0.008$, and $0.005$ for easy, medium, and hard clusters. The interaction coefficients $\eta(d,m)$ are $(.02,.04,.08,.06)$ for easy, $(.04,.06,.06,.05)$ for medium, and $(.08,.06,.03,.03)$ for hard modes ordered as direct, short, long, and code. The length and interaction penalties are clipped at $0.5$ and $0.25$ in absolute value.

\begin{table}[h]
\caption{Mode utility $U(d,m)$ used by MATE-PPO.}
\label{tab:reward-defaults}
\centering
\small
\begin{tabular}{lrrrr}
\toprule
Difficulty & \texttt{direct} & \texttt{short} & \texttt{long} & \texttt{code} \\
\midrule
easy   & $+0.2$ & $+0.1$ & $-0.1$ & $-0.1$ \\
medium & $0.0$  & $+0.2$ & $+0.1$ & $0.0$  \\
hard   & $-0.3$ & $0.0$  & $+0.2$ & $+0.3$ \\
\bottomrule
\end{tabular}
\end{table}

\begin{table}[h]
\caption{Initial reward bands $B^-_{d,m}$--$B^+_{d,m}$ in tokens, computed from the 6,144-token controller scale.}
\label{tab:budget-bands}
\centering
\small
\begin{tabular}{lcccc}
\toprule
Difficulty & \texttt{direct} & \texttt{short} & \texttt{long} & \texttt{code} \\
\midrule
easy   & 16--128 & 96--384  & 384--1536  & 128--768 \\
medium & 32--256 & 160--768 & 768--3072  & 256--1536 \\
hard   & 64--512 & 256--1024 & 1024--4096 & 512--3072 \\
\bottomrule
\end{tabular}
\end{table}

Bands update from correct response lengths after at least 8 observations for a specific key or 12 for a tier-level aggregate. The low and high quantiles are 0.2 and 0.8; maximum downward changes are 5\% and 20\%, and the low target may not exceed 80\% of the high target. Appendix~\ref{app:rollout-clustering} specifies the separate difficulty-clustering state that indexes these bands.

\paragraph{Mode-token parsing and telemetry.}
The completion begins with one generated mode token. The parser maps that token to \texttt{direct}, \texttt{short\_cot}, \texttt{long\_cot}, or \texttt{code}, attaches the selected mode to the rollout record, and removes the control field before task-specific answer extraction. Training uses the token's policy likelihood in the MATE-PPO trajectory; inference aggregates the same field into mode ratios, per-mode accuracy, response length, latency, and budget compliance.

\section{Evaluation Protocol and Reporting Rules}
\label{app:evaluation-rules}

Main-table rows preserve model checkpoint, prompt template, generated-mode protocol, extractor, sample count, decoding parameters, and artifact provenance. All compared models receive the same mode-neutral task prompt and answer-format instruction. For \more{}, the harness parses the generated \texttt{<mode>} token for analysis without conditioning or regenerating the answer. Repeated checkpoints from training runs form the long-horizon trajectories, while component ablations vary one prespecified factor under the same initialization, task stream, update budget, and evaluation protocol.

\subsection{Long-Horizon Checkpoint Selection}
\label{app:longhorizon-checkpoint-selection}
For the paired long-horizon 7B comparisons, the reporting checkpoint for each \more{} variant is selected with the fixed in-distribution score
\[
S_{\mathrm{ID}}(t)=\frac{1}{3}\left[s_{\mathrm{CruxEval-I}}(t)+s_{\mathrm{CruxEval-O}}(t)+s_{\mathrm{LCB\text{-}Execution}}(t)\right],
\]
where each term is the corresponding benchmark score at checkpoint $t$. This rule selects step 250 for both \more{} variants. For the AZR rows in Table~\ref{tab:main-results}, the Code and Math results are retained from the original AZR report, while GSM8K and MMLU-Pro are evaluated at step 350, selected by the same in-distribution score. Thus, the AZR supplemental general-reasoning results use an ID-selected checkpoint, whereas its Code and Math entries retain their original reported protocol.

\subsection{Controlled Ablation Protocol}
\label{app:controlled-ablation-protocol}
Each controlled ablation changes only the component identified by its table row. All shared modules retain identical implementations and configurations, including the backbone initialization, the Stage-1 checkpoint whenever Stage~1 is used, task-generation and rollout settings, task-stream schedule, update budget, decoding settings, mode-neutral evaluation prompts, answer extractors, evaluation cadence, and the in-distribution checkpoint-selection rule above. The \textbf{+10K} controls change only the maximum response length; all remaining shared settings are held fixed.

\subsection{Seed-Averaged Main Results}
\label{app:seed-averaged-main-results}
The two 7B \more{} rows in Table~\ref{tab:main-results} report the mean of three independent training seeds; all non-\more{} rows retain their originally reported values. For model configuration $m$, seed $s\in\{42,0,2026\}$, and any reported aggregate $q\in\{\mathrm{CAvg},\mathrm{MAvg},\mathrm{GSM8K},\mathrm{MMLU\text{-}Pro}\}$, Table~\ref{tab:main-results} reports
\[
\bar q_m=\frac{1}{3}\sum_{s=1}^{3}q_{m,s},
\qquad
\operatorname{SD}(q_m)=\sqrt{\frac{1}{2}\sum_{s=1}^{3}\left(q_{m,s}-\bar q_m\right)^2}.
\]
All values are in percentage points and use the fixed reporting checkpoint from Section~\ref{app:longhorizon-checkpoint-selection}. For \moremodel{}-7B, the standard deviations of CAvg, MAvg, GSM8K, and MMLU-Pro are $0.052$, $0.167$, $0.141$, and $0.108$, respectively; for \moremodel{}-Coder-7B, they are $0.043$, $0.143$, $0.114$, and $0.094$.

\clearpage
\begin{table*}[t]
\section{Scaling Results Across Model Sizes}
\label{app:detailed-scaling-results}

\subsection{Out-of-Distribution Results}
\label{app:scaling-ood-results}

Table~\ref{tab:detailed-scaling-results} reports the full benchmark breakdown for the 3B and 14B scaling experiments summarized in Figure~\ref{fig:scaling-id-ood-results}. CAvg, MAvg, and AVG use the same definitions as Table~\ref{tab:main-results}.

\caption{Detailed out-of-distribution benchmark results for 3B and 14B scaling experiments. CAvg is the mean of \humanevalplus{}, \mbppplus{}, and LiveCodeBench v1--v5; MAvg is the mean of the six mathematical reasoning benchmarks; AVG is the mean of CAvg, MAvg, GSM8K, and M-Pro.}
\label{tab:detailed-scaling-results}
\centering
\small
\setlength{\tabcolsep}{1.0pt}
\renewcommand{\arraystretch}{1.04}
\resizebox{\textwidth}{!}{%
\begin{tabular}{@{}ll|*{4}{c}|*{7}{c}|*{2}{c}|c@{}}
\toprule
\textbf{Model} & \textbf{Variant} & \textbf{\heplus} & \textbf{\mbppplus} & \textbf{LCB} & \textbf{CAvg} & \textbf{A24} & \textbf{A25} & \textbf{AMC} & \textbf{M500} & \textbf{Min.} & \textbf{Oly.} & \textbf{MAvg} & \textbf{GSM8K} & \textbf{M-Pro} & \textbf{AVG} \\
\midrule
Qwen2.5-Coder-3B\tablesource{hui2024qwen25coder} & -- & 67.1 & 65.9 & 20.0 & 51.0 & 3.3 & 3.3 & 20.0 & 51.0 & 18.4 & 16.6 & 18.8 & 69.4 & 38.3 & 44.4 \\
AZR-Coder-3B\tablesource{zhao2025absolutezeroreinforcedselfplay} & +AZR & 71.3 & 69.0 & 24.4 & 54.9 & 3.3 & 3.3 & 37.5 & 62.0 & 26.1 & 27.0 & 26.5 & 71.3 & 40.3 & 48.3 \\
\textsc{MORE}-Coder-3B & +MORE (Ours) & 72.9 & 68.8 & 25.1 & 55.6 & 3.3 & 6.7 & 38.6 & 63.1 & 27.6 & 28.7 & 28.0 & 73.3 & 42.5 & 49.9 \\
\midrule
Qwen2.5-14B-Base\tablesource{qwen2025qwen25technicalreport} & -- & 78.0 & 66.7 & 21.7 & 55.5 & 6.7 & 3.3 & 35.0 & 66.2 & 28.3 & 32.4 & 28.7 & 84.0 & 40.1 & 52.1 \\
AZR-14B\tablesource{zhao2025absolutezeroreinforcedselfplay} & +AZR & 70.7 & 68.8 & 35.2 & 58.2 & 10.0 & 20.0 & 62.5 & 76.2 & 40.4 & 42.5 & 41.9 & 85.2 & 41.9 & 56.8 \\
\textsc{MORE}-14B & +MORE (Ours) & 75.4 & 68.5 & 37.3 & 60.4 & 10.0 & 16.7 & 65.7 & 75.1 & 42.8 & 42.9 & 42.2 & 87.6 & 42.3 & 58.1 \\
\midrule
Qwen2.5-Coder-14B\tablesource{hui2024qwen25coder} & -- & 76.8 & 71.7 & 31.4 & 60.0 & 0.0 & 0.0 & 37.5 & 54.8 & 10.7 & 18.5 & 20.3 & 81.7 & 42.8 & 51.2 \\
AZR-Coder-14B\tablesource{zhao2025absolutezeroreinforcedselfplay} & +AZR & 80.5 & 71.2 & 39.0 & 63.6 & 23.3 & 20.0 & 65.0 & 78.6 & 32.0 & 39.3 & 43.0 & 83.5 & 44.7 & 58.7 \\
\textsc{MORE}-Coder-14B & +MORE (Ours) & 85.4 & 73.3 & 41.5 & 66.7 & 23.3 & 26.7 & 67.5 & 81.2 & 35.5 & 41.6 & 46.0 & 85.1 & 45.8 & 60.9 \\
\bottomrule
\end{tabular}
}
\end{table*}

\subsection{In-Distribution Results}
\label{app:in-distribution-results}

We evaluate the two code-execution task domains used during self-evolution, input prediction and output prediction, with CruxEval-I(nput), CruxEval-O(utput), and LiveCodeBench-Execution (LCB-E) \citep{gu2024cruxeval,jain2024livecodebench}. CruxEval-I is an abduction task: given program behavior, the model recovers an input that produces it. CruxEval-O and LCB-E are deduction tasks, requiring the model to determine program output or execution behavior from a given input and program.

Figure~\ref{fig:scaling-id-ood-results}(a) plots the average of these three metrics throughout training for the Coder-3B, Coder-7B, and Coder-14B settings. The trajectories show an overall increase with occasional intermediate fluctuations, and \more{} traces stronger in-distribution progress than the matched AZR run across all three Coder scales. Because these metrics directly probe the task domains exercised during self-evolution, they provide a targeted measure of in-distribution progress. They complement the out-of-distribution code and mathematical reasoning evaluations reported in Table~\ref{tab:main-results} and Table~\ref{tab:detailed-scaling-results}.

\subsection{Training-Time Efficiency and Difficulty Dynamics}
\label{app:training-time-efficiency}

\begin{figure*}[t]
  \centering
  \includegraphics[width=\textwidth]{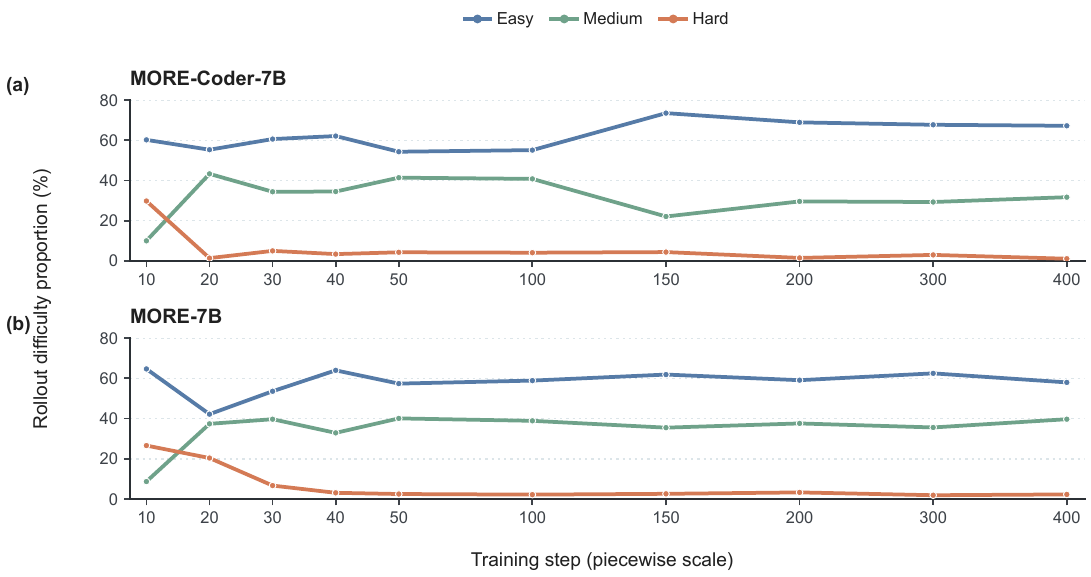}
  \caption{Training-time rollout difficulty-prior proportions. Easy, medium, and hard denote the online difficulty tiers assigned before rollout generation. Allocation is logged every 10 updates through step 50 and at selected 50-step checkpoints thereafter.}
  \label{fig:appendix-difficulty-priors}
\end{figure*}

Figure~\ref{fig:appendix-difficulty-priors} reports the difficulty priors assigned to training rollouts. We log this allocation every 10 updates through step 50. The hard share falls rapidly from its initial level and remains small after step 50: it stays between 1.1\% and 4.4\% for both 7B configurations. Easy and medium tiers retain most of the rollout mass throughout training, but their relative shares continue to shift without a fixed allocation pattern. The later selected 50-step checkpoints retain these meaningful variations after the hard tier has stabilized.

\begin{figure*}[t]
  \centering
  \includegraphics[width=0.78\textwidth]{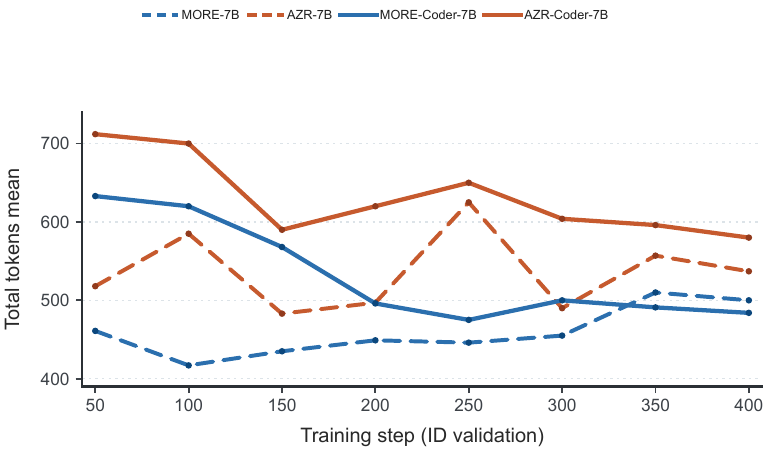}
  \caption{Mean prompt-plus-response token count on the in-distribution validation stream. Base-7B traces are dashed and Coder-7B traces are solid.}
  \label{fig:appendix-id-validation-tokens}
\end{figure*}

Figure~\ref{fig:appendix-id-validation-tokens} evaluates token demand on the in-distribution validation stream. At every measured validation checkpoint, both \moremodel{}-7B and \moremodel{}-Coder-7B have lower mean prompt-plus-response token counts than their matched AZR models. This independent validation-time measurement is consistent with Figure~\ref{fig:training-time-metrics}(b), strengthening the evidence that the lower training rollout demand does not arise only from the sampled training stream.

For phase $\phi\in\{\mathrm{train},\mathrm{val}\}$, we measure GPU-normalized total-token throughput as
\begin{equation}
  \mathrm{total\_tok\_per\_gpu\_s}_{\phi}=
  \frac{\sum_{i\in\mathcal B_{\phi}}\left(|p_i|+|y_i|\right)}{G\,T^{\mathrm{active}}_{\phi}},
  \label{eq:total-tok-per-gpu-second}
\end{equation}
where $p_i$ and $y_i$ denote prompt and response token sequences, $G$ is the visible GPU count, and $T^{\mathrm{active}}_{\phi}$ is the measured active time. Training time includes rollout generation, reward, likelihood, and PPO updates as applicable; validation excludes updates. Checkpointing and cleanup are excluded. Higher values indicate more prompt-plus-response tokens processed per active GPU second.

\begin{figure*}[t]
  \centering
  \includegraphics[width=\textwidth]{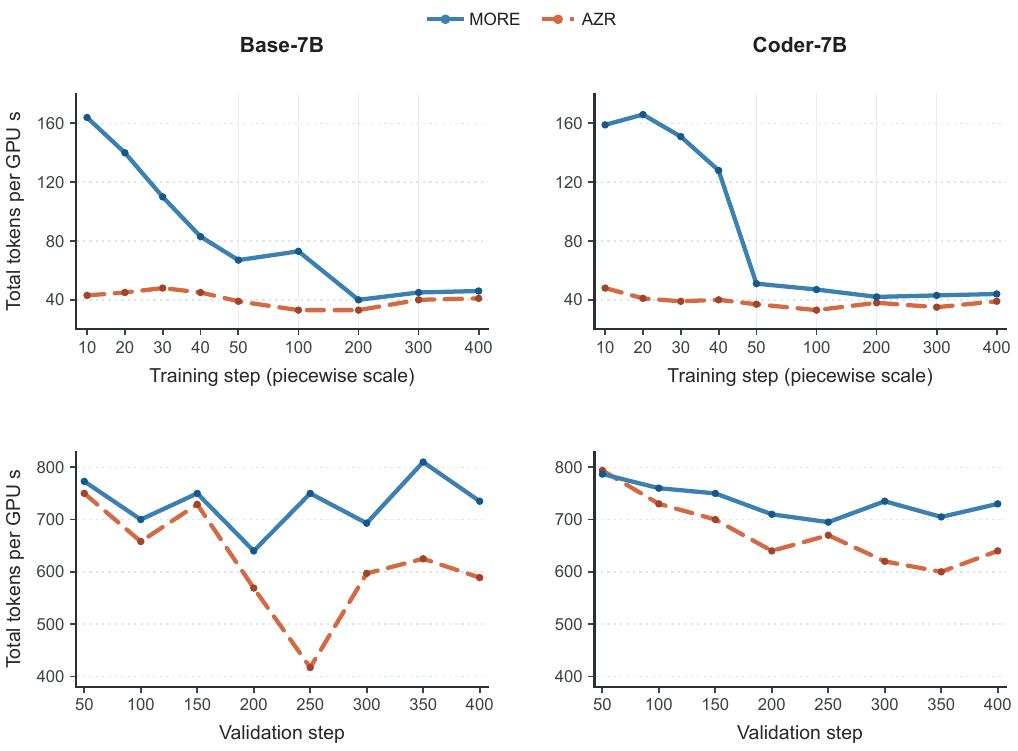}
  \caption{GPU-normalized total-token rate, \texttt{total\_tok\_per\_gpu\_s}, for matched Base-7B and Coder-7B runs. The upper row reports training measurements with the piecewise step scale used in Figure~\ref{fig:training-time-metrics}; the lower row reports equally spaced validation measurements.}
  \label{fig:appendix-gpu-token-rate}
\end{figure*}

Figure~\ref{fig:appendix-gpu-token-rate} reports the full GPU-normalized total-token rate from Equation~\ref{eq:total-tok-per-gpu-second} under matched software, hardware, and hyperparameters. \more{} is higher at every recorded training checkpoint for both Base-7B and Coder-7B. The validation rate is also higher throughout the Base-7B trajectory and at every Coder-7B checkpoint after the initial step-50 measurement; that first Coder validation rate differs by only 7 tokens per GPU second. Taken together with the lower training and validation token demand above, these measurements show a clearly higher training-stage efficiency for \more{} than AZR under the reported protocol.

\section{Research Questions}
\label{app:research-questions}

\subsection{Test-Time Adaptive Reasoning Behavior}
\label{app:test-time-behavior}

\rqbox{1}{How does \more{} distribute its four response contracts across held-out reasoning benchmarks?}

\begin{figure*}[t]
  \centering
  \includegraphics[width=\textwidth]{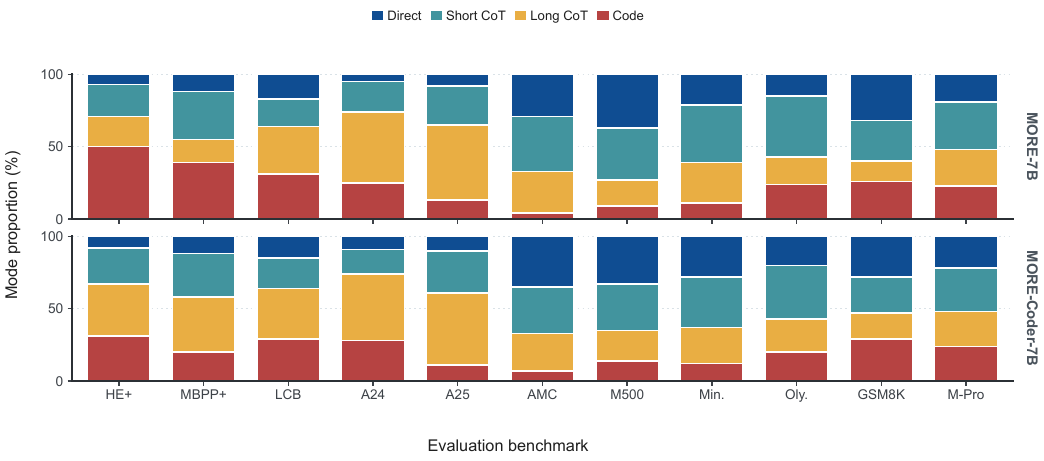}
  \caption{Test-time mode distributions across held-out benchmarks. Each bar gives the percentage of generated mode tokens for one evaluation suite; A24/A25 denote AIME 2024/2025, M500 denotes MATH-500, Min. denotes Minerva Math, Oly. denotes OlympiadBench, and M-Pro denotes MMLU-Pro.}
  \label{fig:appendix-test-time-mode-distribution}
\end{figure*}

Figure~\ref{fig:appendix-test-time-mode-distribution} reports the test-time mode distribution for the two primary 7B \more{} models. All four response contracts contribute on every benchmark, and their proportions vary with task family. \moremodel{}-7B places its largest Code share on \humanevalplus{} and \mbppplus{}, while \moremodel{}-Coder-7B relies on a Long CoT--Code mixture across the three coding benchmarks. Both models raise their Long CoT allocation markedly on AIME 2024 and AIME 2025. The remaining mathematical and general-reasoning evaluations are handled through distinct Direct, Short CoT, and Long CoT mixtures, showing task-specific allocation of reasoning form across the full evaluation suite.

\begin{figure*}[t]
  \centering
  \includegraphics[width=\textwidth]{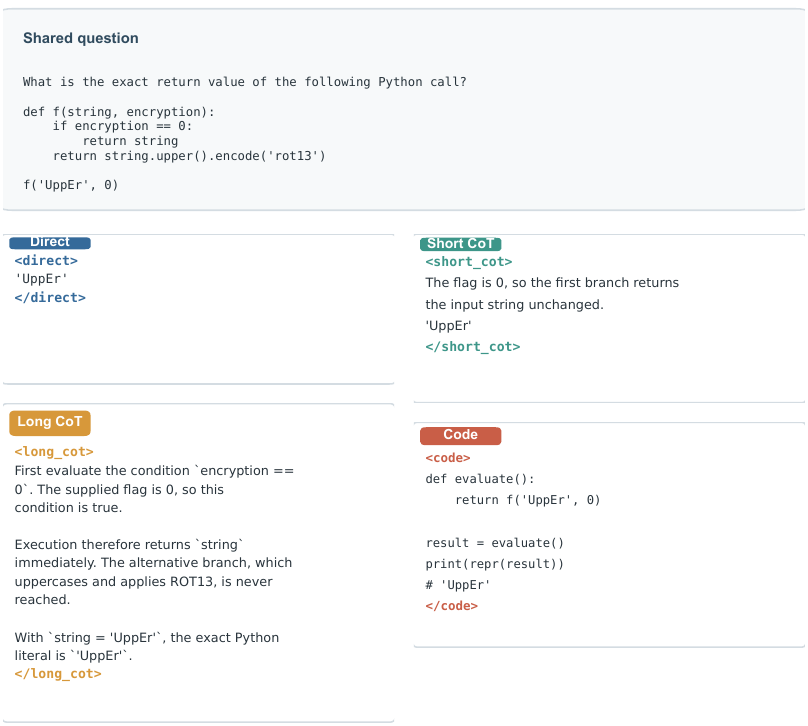}
  \caption{\textbf{CruxEval-O (sample 8): Four-mode response contracts.} A shared program-output task is expressed as an answer-only response, a compact condition check, an expanded execution trace, and an executable evaluation.}
  \label{fig:appendix-mode-example-code}
\end{figure*}

\begin{figure*}[t]
  \centering
  \includegraphics[width=\textwidth]{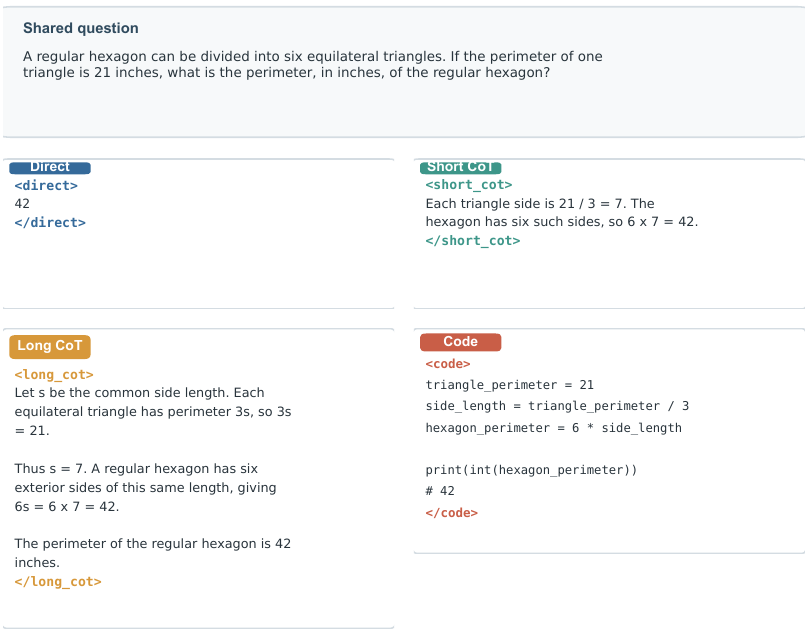}
  \caption{\textbf{MATH-500 (prealgebra problem 1622): Four-mode response contracts.} A shared geometry task is expressed through an answer-only response, a compact numerical derivation, an expanded symbolic derivation, and an executable calculation.}
  \label{fig:appendix-mode-example-math}
\end{figure*}

Figures~\ref{fig:appendix-mode-example-code} and~\ref{fig:appendix-mode-example-math} make the four response contracts concrete on a common code task and a common mathematical task. Direct isolates the final answer, Short CoT retains only the decisive inference, Long CoT develops the governing relation through explicit intermediate steps, and Code expresses the same solution as an executable procedure. On the program-output task, the contracts range from a branch decision to an explicit evaluation trace; on the geometry task, they range from a compact perimeter calculation to a symbolic derivation and a short program. The contrast makes the four response forms readily identifiable while preserving a common target answer.

\clearpage

\subsection{Test-Time Response-Efficiency Patterns}
\label{app:test-time-efficiency}

\rqbox{2}{Does adaptive mode selection achieve a stronger test-time accuracy--efficiency trade-off than a fixed reasoning mode?}

To isolate the value of routing, Table~\ref{tab:fixed-mode-routing} compares the adaptive \moremodel{}-Coder-7B policy with four counterparts that apply one response contract to every evaluation instance. Each benchmark pair reports accuracy and Response Length Avg (RLA), making the capability--effort profile of a globally fixed mode directly comparable with adaptive routing.

\begin{table*}[t]
  \caption{Fixed-mode and adaptive evaluation of \moremodel{}-Coder-7B. Fixed policies apply the indicated reasoning contract to every instance. Accuracy is reported in percentage points; Response Length Avg (RLA) is the mean number of generated response tokens, where lower is more efficient. The best and second-best accuracy in each suite are shown in \textbf{bold} and \underline{underlined}, respectively.}
  \label{tab:fixed-mode-routing}
  \centering
  \small
  \setlength{\tabcolsep}{5pt}
  \renewcommand{\arraystretch}{1.10}
  \begin{tabular}{lcc|cc|cc|cc}
    \toprule
    \textbf{Policy} & \multicolumn{2}{c|}{\textbf{Code Avg}} & \multicolumn{2}{c|}{\textbf{Math Avg}} & \multicolumn{2}{c|}{\textbf{GSM8K}} & \multicolumn{2}{c}{\textbf{MMLU-Pro}} \\
    & \textbf{Acc.} $\uparrow$ & \textbf{RLA} $\downarrow$ & \textbf{Acc.} $\uparrow$ & \textbf{RLA} $\downarrow$ & \textbf{Acc.} $\uparrow$ & \textbf{RLA} $\downarrow$ & \textbf{Acc.} $\uparrow$ & \textbf{RLA} $\downarrow$ \\
    \midrule
    Fixed \texttt{direct}    & 49.4 & 77.4   & 33.7 & 537.5  & 60.7 & 43.8  & 37.5 & 153.3 \\
    Fixed \texttt{short\_cot} & 51.2 & 103.7  & 37.4 & 1,216.8 & 67.4 & 87.7  & 43.2 & 254.6 \\
    Fixed \texttt{long\_cot}  & \underline{61.6} & 405.8  & \textbf{43.1} & 4,762.3 & \textbf{79.1} & 122.4 & \textbf{46.9} & 435.7 \\
    Fixed \texttt{code}      & 55.3 & 156.5  & 39.2 & 2,066.1 & 70.9 & 91.6  & 41.3 & 269.0 \\
    \midrule
    \rowcolor{black!4}\textbf{Adaptive \moremodel{}-Coder-7B} & \textbf{62.8} & 191.7 & \underline{42.8} & 3,015.9 & \underline{77.8} & 105.1 & \underline{46.0} & 303.4 \\
    \bottomrule
  \end{tabular}
\end{table*}

The fixed policies expose a clear capability--effort ladder: Direct yields the shortest responses, while Long CoT achieves the strongest fixed-policy accuracy at substantially greater token cost. Adaptive routing attains the highest Code Avg of all policies while using 52.8\% fewer Code tokens than fixed Long CoT. On Math Avg, GSM8K, and MMLU-Pro, it remains within 0.3, 1.3, and 0.9 points of the strongest fixed policy, respectively, while reducing RLA by 36.7\%, 14.1\%, and 30.4\%. The learned policy thus draws on the strengths of long reasoning when it is useful without committing every instance to its cost, yielding the strongest observed accuracy--efficiency balance across the evaluated suites.

The benchmark-specific allocations in Figure~\ref{fig:appendix-test-time-mode-distribution} align with the compact test-time profiles in Table~\ref{tab:ood-evaluation-efficiency}. Across both Base and Coder model families, \more{} couples these varied mode mixtures with lower Response Length Avg and Wall Time than matched AZR models. The resulting test-time behavior therefore combines task-specific response-form selection with lower overall generation demand and elapsed evaluation time.

\subsection{Interpretable Terminal-Return Dynamics}
\label{app:terminal-return-dynamics}

\rqbox{3}{How do the components of the \more{} terminal return evolve during self-evolution?}

Across steps 100--400, $R_{\mathrm{MF}}$ stays positive and comparatively stable, whereas $R_{\mathrm{acc}}$ changes with stream-level outcomes; $R_{\mathrm{BD}}$ and $R_{\mathrm{MB}}$ remain small corrections. The raw and clipped $R_{\mathrm{BD}}$ traces coincide throughout the recorded trajectory, indicating that the observed mean deviation stays within the active clipping interval. Thus, MATE-PPO combines a task-dependent correctness signal with a persistent mode-fit preference while keeping budget regulation bounded and non-dominant.

\begin{figure*}[t]
  \centering
  \includegraphics[width=0.94\textwidth]{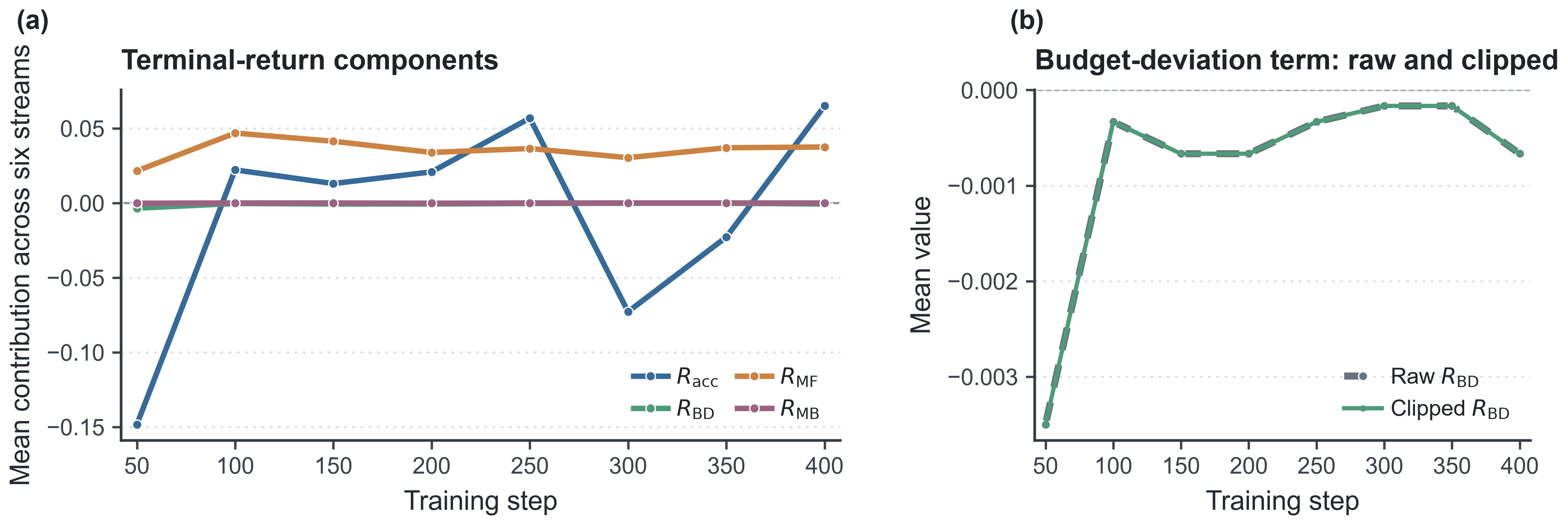}
  \caption{Terminal-return telemetry for the full \moremodel{}-Coder-7B trajectory. Values at steps 50--400 are means across the six proposer--solver streams. (a) Contributions of $R_{\mathrm{acc}}$, $R_{\mathrm{BD}}$, $R_{\mathrm{MF}}$, and $R_{\mathrm{MB}}$. (b) Raw and clipped budget-deviation telemetry; their overlap denotes values within the active clipping interval.}
  \label{fig:appendix-terminal-return-components}
\end{figure*}

\subsection{Self-Generated Task-Stream Diversity}
\label{app:task-stream-diversity}

\rqbox{4}{Does \more{} sustain a structurally varied self-generated task stream throughout training?}

From step 100 onward, output-generator complexity remains above input-generator complexity, both edit-distance traces stay near $0.6$, and answer diversity remains high across the input, output, and function pathways. These separated structural trajectories and persistently high diversity show that the construction pathways do not converge to a shared task template. \more{} therefore sustains a structurally varied self-generated task stream throughout the recorded training horizon.

\begin{figure*}[t]
  \centering
  \includegraphics[width=0.94\textwidth]{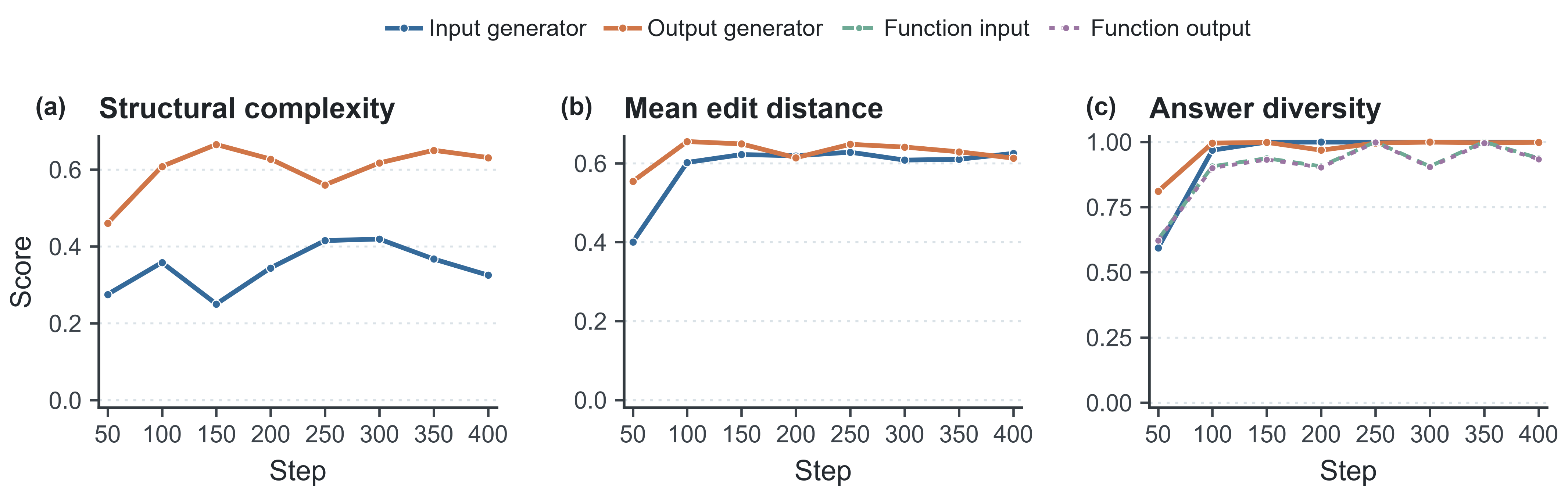}
  \caption{Task-construction proxies for the full \moremodel{}-Coder-7B trajectory. Structural complexity and mean edit distance are reported for the input and output generators; answer diversity additionally reports the function stream's input- and output-answer diversity.}
  \label{fig:appendix-task-stream-diversity}
\end{figure*}

\subsection{Mode-Conditioned Budget Calibration}
\label{app:mode-budget-calibration}

\rqbox{5}{Do MATE-PPO's mode-conditioned budget bands calibrate realized effort across the four reasoning contracts?}

After the initial update, Direct, Short CoT, Long CoT, and Code occupy distinct dynamic $B^-$--$B^+$ scales, with Direct lowest and the longer reasoning forms assigned higher bands. From step 100 onward, the in-band fraction remains between $53\%$ and $65\%$, while the mode-wise under- and over-band deviations stay bounded around the corresponding moving targets. These trajectories reflect stochastic rollout-level variation around mode-specific bands rather than a single static generation ceiling. MATE-PPO therefore calibrates realized effort against separate, evolving targets for each reasoning contract.

\begin{figure*}[t]
  \centering
  \includegraphics[width=0.96\textwidth]{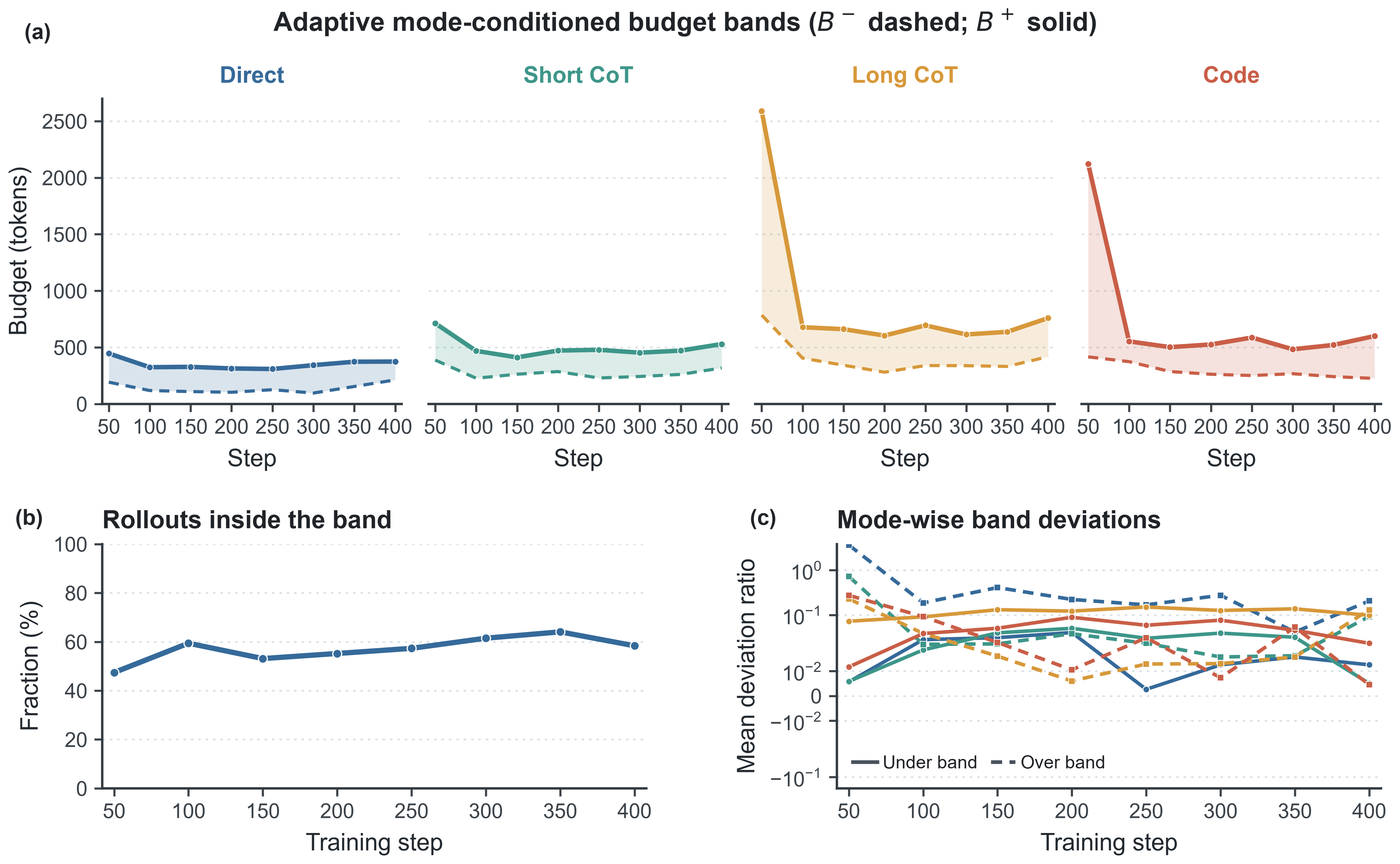}
  \caption{Mode-conditioned budget calibration for the full \moremodel{}-Coder-7B trajectory. (a) Dynamic $B^-$ (dashed) and $B^+$ (solid) bands for each reasoning mode. (b) Fraction of rollouts within the active band. (c) Mode-wise under-band and over-band deviation ratios. All traces are means across the six proposer--solver streams at steps 50--400.}
  \label{fig:appendix-budget-band-calibration}
\end{figure*}

\subsection{Role-Sensitive Mode Utility}
\label{app:stream-mode-utility}

\rqbox{6}{Does \more{} allocate modes differently across proposer--solver streams in ways that are interpretable through verifier feedback?}

Input streams favor Direct and Short CoT, and the output proposer likewise favors Short CoT; the output solver and function streams place more mass on Long CoT. Verifier-success profiles also vary by stream, with the strongest observed mode switching between Direct and Short CoT across task families and roles. \more{} therefore learns role-sensitive mode mixtures rather than collapsing to a universal fixed-mode policy.

\begin{figure*}[t]
  \centering
  \includegraphics[width=0.88\textwidth]{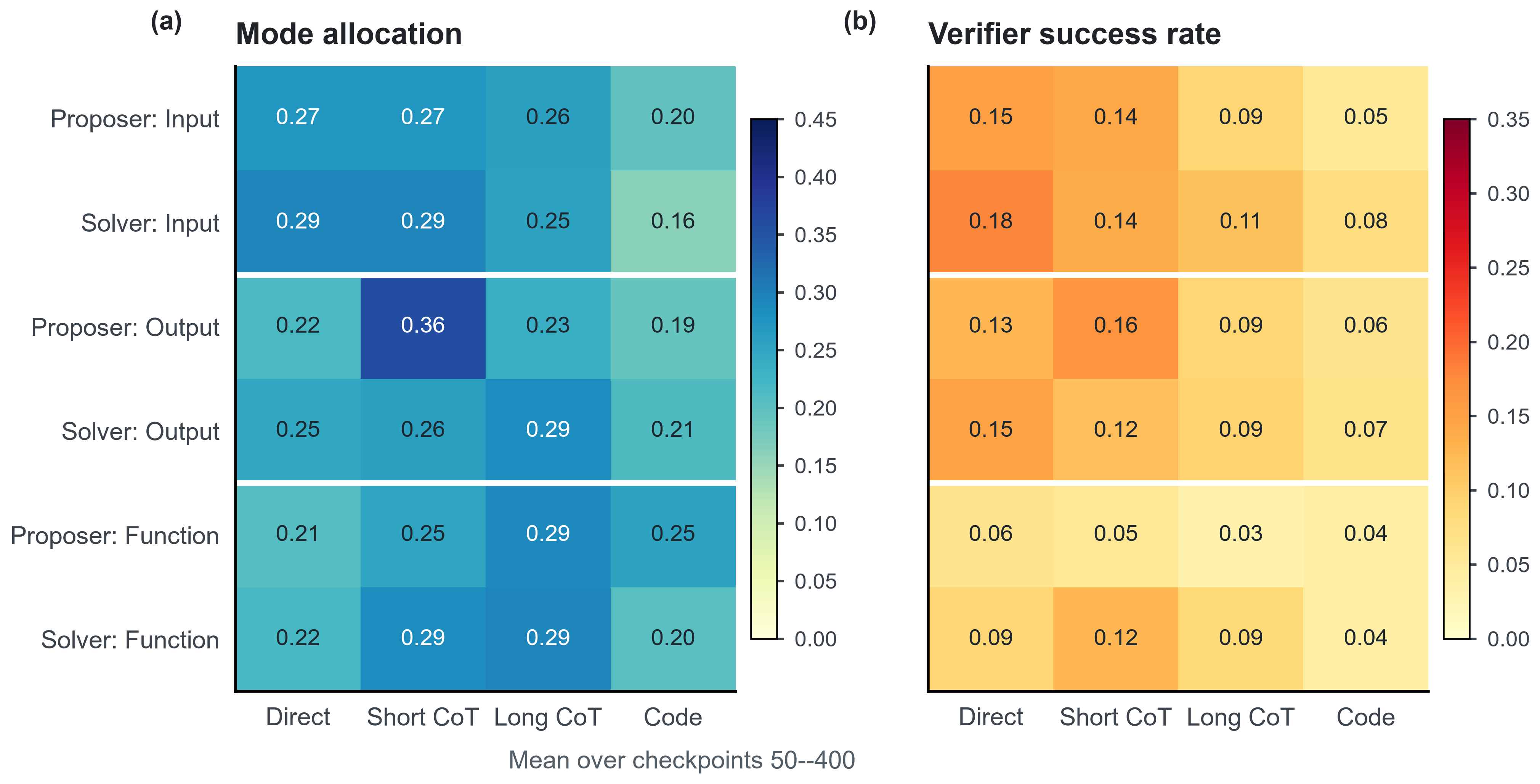}
  \caption{Mode allocation and verifier success by proposer--solver stream for \moremodel{}-Coder-7B. Each cell is the mean value over checkpoints 50--400. The left matrix gives the allocation ratio of each reasoning mode; the right matrix gives the corresponding verifier success rate.}
  \label{fig:appendix-stream-mode-utility}
\end{figure*}

\end{document}